\documentclass[letterpaper,twocolumn,10pt]{article}
\usepackage{usenix}
\usepackage{tikz,pifont,algorithm,algorithmic,amssymb,tabularx,multirow,amsmath,comment,threeparttable,booktabs,adjustbox, caption,colortbl,bbm,makecell,hyperref,xspace,xcolor}
\usepackage[T1]{fontenc}
\definecolor{mygreen}{rgb}{0,0.7,0}
\newcommand{\system}{{\sc Unidoor}\xspace}

\begin{document}

\date{}

\title{\system: A Universal Framework for Action-Level Backdoor Attacks in Deep Reinforcement Learning}

\author{
{\rm Oubo Ma}\\
Zhejiang University
\and
{\rm Linkang Du}\\
Xi'an Jiaotong University
\and
{\rm Yang Dai}\\
Laboratory for Big Data and Decision
\and
{\rm Chunyi Zhou}\\
Zhejiang University
\and
{\rm Qingming Li}\\
Zhejiang University
\and
{\rm Yuwen Pu}\\
Zhejiang University
\and
{\rm Shouling Ji*}\\
Zhejiang University
} 

\maketitle

\newcommand\blfootnote[1]{%
	\begingroup
	\renewcommand\thefootnote{}\footnote{#1}%
	\addtocounter{footnote}{-1}%
	\endgroup
}

\begin{abstract}
  Deep reinforcement learning (DRL) is widely applied to safety-critical decision-making scenarios. 
  However, DRL is vulnerable to backdoor attacks, especially action-level backdoors, which pose significant threats through precise manipulation and flexible activation, risking outcomes like vehicle collisions or drone crashes.
  The key distinction of action-level backdoors lies in the utilization of the backdoor reward function to associate triggers with target actions.
  Nevertheless, existing studies typically rely on backdoor reward functions with fixed values or conditional flipping, which lack universality across diverse DRL tasks and backdoor designs, resulting in fluctuations or even failure in practice.

  This paper proposes the first universal action-level backdoor attack framework, called \system, which enables adaptive exploration of backdoor reward functions through performance monitoring, eliminating the reliance on expert knowledge and grid search.
  We highlight that action tampering serves as a crucial component of action-level backdoor attacks in continuous action scenarios, as it addresses attack failures caused by low-frequency target actions.
  Extensive evaluations demonstrate that \system significantly enhances the attack performance of action-level backdoors, showcasing its universality across diverse attack scenarios, including single/multiple agents, single/multiple backdoors, discrete/continuous action spaces, and sparse/dense reward signals.
  Furthermore, visualization results encompassing state distribution, neuron activation, and animations demonstrate the stealthiness of \system.
  The source code of \system can be found at \href{https://github.com/maoubo/UNIDOOR}{https://github.com/maoubo/UNIDOOR}.
\end{abstract}
\blfootnote{*Corresponding author.}
\section{Introduction}
\label{sec:introduction}

Deep reinforcement learning (DRL) has achieved significant milestones, including superhuman-level Go AI~\cite{silver2017mastering}, protein structure prediction~\cite{jumper2021highly}, matrix multiplication discovery~\cite{fawzi2022discovering}, and safety alignment for large language models~\cite{achiam2023gpt}.
However, DRL faces substantial threats, such as backdoor attacks, where malicious backdoors are injected into the victim agent's policy during training and activated via triggers during deployment.

Current backdoor attacks against DRL are categorized into two types: policy-level and action-level.
Policy-level backdoors~\cite{yang2019design,wang2021backdoorl,gong2024baffle} manipulate the victim agent's long-term objectives, with each trigger corresponding to a target policy. 
However, they lack precise control over the actions and require retraining when the attack objective evolves.
In contrast, action-level backdoors~\cite{kiourti2020trojdrl,ashcraft2021poisoning,chen2022marnet,cui2024badrl,rathbun2024sleepernets} emphasize precise control over the victim agent's step-by-step actions. 
This enables the adversary to flexibly activate the backdoor to align with various attack objectives, posing a heightened threat to safety-critical applications.
We conduct a comprehensive comparison of the two types of backdoors in Section~\ref{sec:backdoor attacks}, accompanied by examples.

Existing action-level backdoor attacks predominantly draw on the poisoning paradigm from deep learning (DL)~\cite{shen2021backdoor,carlini2022poisoning,lv2023data}.
The key distinction lies in the adversary targeting transitions (state, action, reward triplets) in DRL rather than sample-label pairs in DL. 
In this case, a trigger is embedded into the state and bound to a target action via the backdoor reward function.
However, these studies prioritize trigger design and target action selection while defining backdoor reward functions in simplistic forms, such as fixed values~\cite{kiourti2020trojdrl} or conditional flipping~\cite{ashcraft2021poisoning}, predetermined by the adversary on a case-by-case basis.
Through an empirical study, we reveal that the case-by-case approach limits the universality of current action-level backdoor attacks in practical applications, resulting in performance fluctuations or outright failure.

\begin{figure}
    \centering
    \includegraphics[width=0.5\textwidth]{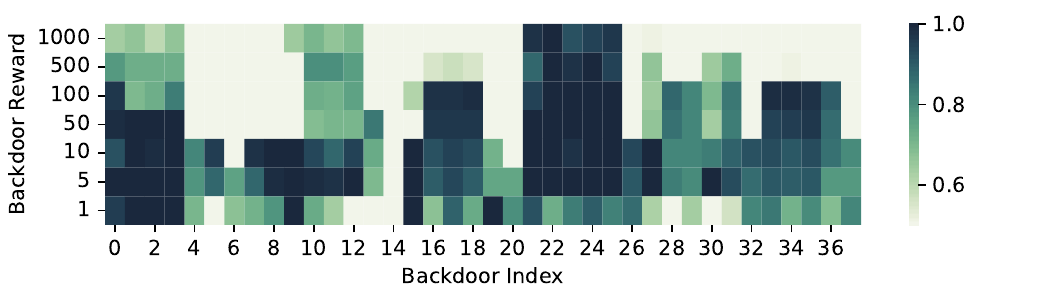}
    \caption{The impact of backdoor reward and task variations on the attack performance of action-level backdoors, where color intensity reflects the harmonic mean of the victim agent's performance on both benign and backdoor tasks.}
    \label{fig:empirical_study}
\end{figure}

\noindent \textbf{Empirical Study.}
We take TrojDRL~\cite{kiourti2020trojdrl} as an example, where the backdoor reward is defined as a fixed value, to investigate how variations in backdoor reward and task settings affect attack performance.
We select 6 DRL tasks from OpenAI Gym~\cite{gym} and design 38 action-level backdoor tasks (indices 0-37 in Table~\ref{tab:backdoor_design}).
The results in Figure~\ref{fig:empirical_study} indicate that:
(1) The attack performance exhibits fluctuations or outright failure in response to changes in the backdoor reward.
(2) It is infeasible to predetermine a static backdoor reward that achieves universality across all benign and backdoor tasks.

\noindent \textbf{Motivation.}
Based on the aforementioned observations, we argue that the non-stationarity of DRL training and the variability of benign reward functions across tasks restrict the universality of existing action-level backdoor attacks.
Additionally, the manual design of backdoor reward functions relies on expert knowledge and extensive trial-and-error, increasing attack complexity and entry barriers.
Therefore, this paper aims to (1) achieve the adaptive adjustment of the backdoor reward function to enhance the universality of action-level backdoor attacks, and (2) reveal potential vulnerabilities in DRL and highlight its security implications to foster greater awareness and attention within the research community.

\noindent \textbf{Challenges.}
We encounter three main challenges:
(1) \textit{Task Discrepancy}.
The differences in Markov Decision Process (MDP) modeling across tasks present distinct challenges for backdoor injection.
(2) \textit{Distraction Dilemma}.
Simultaneously training benign and backdoor tasks within a shared policy in DRL may result in one task dominating the process.
(3) \textit{Limited Trial-and-Error Search}.
DRL's inherent non-stationarity necessitates reducing the frequency of adjustments to the backdoor reward function to avoid performance instability or irrecoverable degradation in both benign and backdoor tasks.

\noindent \textbf{Our Proposal.}
This paper introduces \system, a universal framework that conceptualizes action-level backdoor attacks as a multi-task learning paradigm, comprising four main modules: \textit{Performance Monitoring}, \textit{Initial Freezing}, \textit{Transition Poisoning}, and \textit{Adaptive Exploration}.
\textit{Performance Monitoring} tracks and normalizes the performance of the victim agent on both benign and backdoor tasks based on trajectories and transitions. 
This task-agnostic approach mitigates \textit{Task Discrepancy} (Challenge 1).
\textit{Initial Freezing} introduces a stabilization phase to avoid the dominance of backdoor tasks during early training stages, thereby effectively mitigating \textit{Distraction Dilemma} (Challenge 2).
After freezing ends, \textit{Transition Poisoning} is used to inject action-level backdoors, incorporating action tampering to support both discrete and continuous action scenarios.
Simultaneously, \textit{Adaptive Exploration} is activated to adjust the backdoor reward adaptively.
It integrates monitored performance to conduct conservative exploration, reducing frequent adjustments and addressing \textit{Limited Trial-and-Error Search} (Challenge 3).

\noindent \textbf{Evaluations.}
Extensive evaluations demonstrate the effectiveness of \system across 11 DRL tasks, 53 backdoor designs, and 3 mainstream DRL algorithms, covering diverse attack scenarios, including single/multiple agents, single/multiple backdoors, discrete/continuous action spaces, and sparse/dense reward signals (see Table~\ref{tab:tasks} and Table~\ref{tab:backdoor_design}).
Visualization evaluations show the stealthiness of \system via state distribution, neuron activation, and visual animations.
Moreover, we explore two potential defenses and reveal that the inherent instability of DRL renders traditional DL defense strategies ineffective against action-level backdoor attacks.

\noindent \textbf{Contributions. }
In summary, the main contributions of this paper are fourfold:
\begin{itemize}
  \item To the best of our knowledge, \system is the first universal framework for action-level backdoor attacks, revealing that the adversary can achieve cross-task backdoor injection without relying on expert knowledge or extensive trial-and-error.
  \item Building on our finding that benign task performance inversely correlates with backdoor reward while backdoor task performance positively correlates, we implement adaptive exploration of the backdoor reward function to enhance the universality of \system.
  \item We highlight that action tampering is a crucial component of action-level backdoor attacks in continuous action scenarios, as it addresses attack failures caused by low-frequency target actions.
  \item We conduct a systematic evaluation, demonstrating that \system outperforms state-of-the-art methods in diverse scenarios and the adversary can flexibly design activation strategies to achieve specific attack objectives using action-level backdoors.
\end{itemize}

\section{Background}
\label{sec:background}

\subsection{Deep Reinforcement Learning}
\label{sec:drl}

DRL is a machine learning paradigm in which an agent learns to make optimal sequential decisions within an environment by maximizing cumulative rewards through trial and error.
This process is modeled as a MDP, which represented as $\mathcal{M} = (\mathcal{S}, \mathcal{A}, \mathcal{R}, \mathcal{P}, \gamma)$, where $\mathcal{S}$ is the state space.
$\mathcal{A}$ is the action space.
$\mathcal{R}:\mathcal{S} \times \mathcal{A} \rightarrow \mathbb{R}$ is the reward function, indicating the immediate reward that the agent receives from the environment for taking action $a \in \mathcal{A}$ in state $s \in \mathcal{S}$.
$\mathcal{P}:\mathcal{S} \times \mathcal{A} \rightarrow \Delta (\mathcal{S})$ is the state transition function, indicating the probability that taking action $a \in \mathcal{A}$ in state $s \in \mathcal{S}$ results in a transition to $s' \in \mathcal{S}$.
$\gamma \in [0,1)$ is the discount rate, which determines the present value of future rewards.

\begin{figure}
    \centering
    \includegraphics[width=0.4\textwidth]{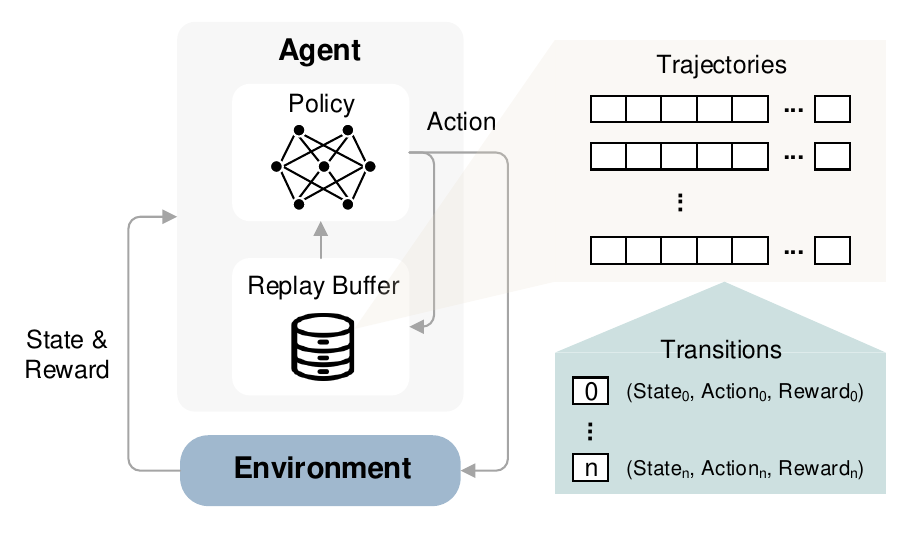}
    \caption{The schematic diagram of DRL.}
    \label{fig:drl}
\end{figure}

The agent makes decisions based on the deep neural network policy $\pi_\theta$, where $\pi_\theta:\mathcal{S} \rightarrow \Delta (\mathcal{A})$ maps state $s \in \mathcal{S}$ to a specific action or a probability distribution over actions in $\mathcal{A}$.
As shown in Figure~\ref{fig:drl}, the agent stores interaction experiences with the environment from an episode as a trajectory $\tau = \{ \tau_0, \tau_1, ..., \tau_T \}$, where $\tau_t = (s_t, a_t, r_t)$ denotes the transition of time step $t$, $T$ is the time horizon, and $r$ is the immediate reward.
The goal of DRL is to find the optimal parameters $\theta$ that maximize the expected cumulative reward over time,
\begin{equation}
\theta^* = \mathop{\arg\max}\limits_{\theta} \mathbb{E}_{\tau \sim \pi_\theta} [\mathcal{G}(\tau)],
\label{eq:rl}
\end{equation}
where $\mathcal{G}(\tau) = \sum_{t=0}^{T} \gamma^t r_t$ is the discounted return.

\subsection{DRL Backdoor Attacks}
\label{sec:backdoor attacks}

DRL backdoor attacks aim to force the victim agent to execute a specific policy or action, and they can be categorized into the following two types. 

\noindent \textbf{Policy-Level Backdoor.}
This type of backdoor focuses on coarse-grained manipulation of the victim agent, with each trigger mapping to a target policy.
For instance, the adversary activates the backdoor to force an autonomous vehicle to redirect its destination from a school to a hospital, regardless of the specific route taken.
Yang \emph{et al.}~\cite{yang2019design} proposed an environment switching framework, while Wang \emph{et al.}~\cite{wang2021backdoorl} introduced a policy combination approach guided by behavior cloning to inject policy-level backdoors.
Gong \emph{et al.}~\cite{gong2024baffle} proposed generating poisoned trajectories from a backdoor policy to implement policy-level backdoor attacks in offline RL.

\noindent \textbf{Action-Level Backdoor.}
This type of backdoor focuses on fine-grained manipulation of the victim agent, with each trigger mapping to a target action.
For instance, an adversary might activate the backdoor at a critical moment to force an autonomous driving agent to execute a sharp turn, potentially causing traffic congestion or collisions.
Kiourti \emph{et al.}~\cite{kiourti2020trojdrl} proposed a transition poisoning paradigm for injecting action-level backdoors.
Building on this approach, Ashcraft \emph{et al.}~\cite{ashcraft2021poisoning} enhanced stealthiness using in-distribution triggers, while Cui \emph{et al.}~\cite{cui2024badrl} improved effectiveness through state-based trigger optimization.
These methods employ predefined static backdoor rewards, such as constants, conditional flipping, or the minimum positive reward signal, disregarding the adverse effects of policy dynamics on the attack.
To address this, Chen \emph{et al.}~\cite{chen2022marnet} proposed a dynamic backdoor reward mechanism, while Rathbun \emph{et al.}~\cite{rathbun2024sleepernets} utilized Monte Carlo estimation to ensure the target action is theoretically optimal.
However, these methods remain task-specific, require expert knowledge, and lack universality across benign and backdoor tasks, posing a considerable entry barrier.

\noindent \textbf{Backdoor Comparison.}
Policy-level backdoors excel in exerting coarse-grained influence on the victim agent's long-term decisions per trigger, yet their limitations are apparent:
(1) Controlling most actions through a few triggers sacrifices precision, such as the duration of the backdoor activation and the victim agent's action output per time step.
(2) The victim agent requires modules like LSTM to process sequential data and retain temporal information.
(3) Attack techniques such as environment switching and policy combination require training privileges and control over the victim agent's training schedule.
(4) The adversary must synchronously update the backdoor policy whenever the attack objective evolves.

Conversely, action-level backdoors exhibit a short-term activation nature.
The one-to-one correspondence between triggers and target actions, independent of the model structure, allows for fine-grained and precise backdoor activation.
This allows the adversary to selectively activate or terminate the backdoor at specific time steps and accurately control the victim agent's actions.
As these backdoors are primarily injected through the poisoning paradigm, the adversary typically does not require training privileges for the victim agent.
Furthermore, the adversary can change the activation strategy to switch attack objectives.
For instance, redirecting the victim agent's destination from a hospital to a supermarket necessitates retraining in the case of policy-level backdoors. 
In contrast, action-level backdoors require only the design of a new activation strategy, leveraging repeated triggers to achieve the desired redirection.

We summarize the distinctions between policy-level and action-level backdoor attacks across 6 criteria in Table~\ref{tab:comparison_table}.
Based on the comparison, this paper focuses on action-level backdoor attacks for their precise manipulation and flexible activation. 
It addresses the reliance on expert knowledge and the lack of universality in existing methods.

\section{Threat Model}
\label{sec:threat model}

In this paper, the attack scenario involves two parties: the victim and the adversary.
The victim employs a DRL algorithm to train a policy $\pi_\theta$ for a benign task $\mathcal{M} = (\mathcal{S}, \mathcal{A}, \mathcal{R}, \mathcal{P}, \gamma)$.
The adversary aims to inject action-level backdoors into the policy $\pi_\theta$.
The backdoor task is defined by the tuple $(\mathcal{T}, \mathcal{S}^\dag, \mathcal{A}^\dag, \mathcal{F}_s, \mathcal{F}_a, \mathcal{R}^\dag)$, where $\mathcal{T}$ is the trigger space.
$\mathcal{S}^\dag \subseteq \mathcal{S}$ is the subset of states containing embedded triggers.
$\mathcal{A}^\dag \subseteq \mathcal{A}$ is the target action space.
$\mathcal{F}_s:\mathcal{S} \times \mathcal{T} \rightarrow \mathcal{S}^\dag$ is the trigger-state mapping function, defining how a state is transformed when a trigger is embedded into it.
$\mathcal{F}_a:\mathcal{T} \rightarrow \mathcal{A}^\dag$ is the trigger-action mapping function, establishing a bijection (both injective and surjective) between triggers and target actions.
This ensures a one-to-one correspondence, i.e., $|\mathcal{T}| = |\mathcal{A}^\dag|$, meaning that the cardinality of $\mathcal{T}$ and $\mathcal{A}^\dag$ equal\footnote{This setup is designed to facilitate formalization and understanding. 
In practical attack scenarios, the adversary can define multiple triggers corresponding to a single target action, i.e., $|\mathcal{T}| \geq |\mathcal{A}^\dag|$.}.
$\mathcal{R}^\dag$ is the backdoor reward function that establishes and reinforces the binding relationship between triggers and their corresponding target actions.

\noindent \textbf{Adversary's Objective.}
The backdoored policy is defined as $\pi_{\theta^\dag}$.
If $\mathcal{A}$ is a discrete space, the adversary aims for $\pi_{\theta^\dag}$ to output the target action with the highest probability whenever the input state contains a trigger $\delta \in \mathcal{T}$, formally defined as
\begin{equation}
\max_{\theta^\dag} \mathbb{E}_{s \sim \mathcal{S}, \delta \sim \mathcal{T}} \left[ \pi_{\theta^\dag}(a^\dag | \mathcal{F}_s (s, \delta)) \right], \text{where } a^\dag = \mathcal{F}_a (\delta).
\label{eq:form_dis}
\end{equation}
If $\mathcal{A}$ is a continuous space, the adversary aims for $\pi_{\theta^\dag}$ to output actions that minimize the distance to the target action when the input state contains a trigger, formally defined as
\begin{equation}
\min_{\theta^\dag} \mathbb{E}_{s \sim \mathcal{S}, \delta \sim \mathcal{T}} \left[ || \pi_{\theta^\dag} (\mathcal{F}_s (s, \delta)) - a^\dag || \right], \text{where } a^\dag = \mathcal{F}_a (\delta).
\label{eq:form_con}
\end{equation}
To ensure stealthiness, the adversary aims for $\pi_{\theta^\dag}$ to produce sequential decision-making indistinguishable from $\pi_\theta$ in the absence of a trigger in the input state, formally defined as
\begin{equation}
\min_{\theta^\dag} \mathbb{E}_{s \sim \mathcal{S} \setminus \mathcal{S}^\dag} \left[ D(\pi_{\theta^\dag} (\cdot | s), \pi_\theta (\cdot | s)) \right],
\label{eq:stealthiness}
\end{equation}
where $D(\cdot, \cdot)$ is a divergence metric (e.g., Kullback-Leibler divergence~\cite{kullback1951information}) that measures the difference between the action probability distributions of $\pi_{\theta^\dag}$ and $\pi_\theta$ given state $s$.

\noindent \textbf{Adversary's Capability.}
The adversary is able to perturb the victim's observations and access and modify the transitions in the victim's replay buffer.
This assumption is less restrictive than those in existing works~\cite{kiourti2020trojdrl,ashcraft2021poisoning,chen2022marnet,cui2024badrl,rathbun2024sleepernets}, as it imposes no specific requirements on the victim's training privileges, training schedule, DRL algorithm, model structure, or hyperparameter settings.
For clarity, we provide several concrete attack scenarios in Appendix~\ref{app:attack_scenarios}.

\begin{figure*}
    \centering
    \includegraphics[width=0.85\textwidth]{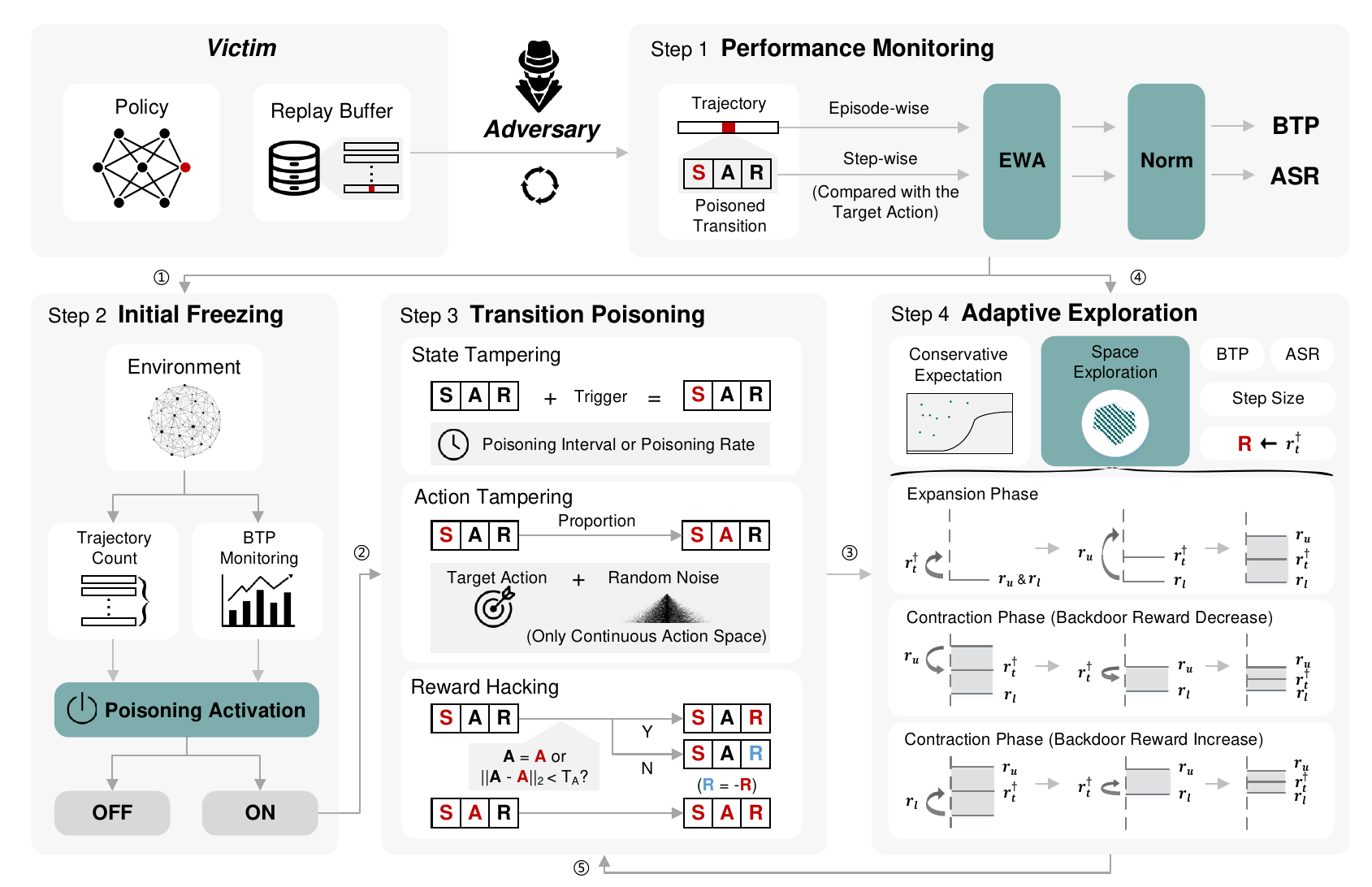}
    \caption{The framework of \system. \ding{172} is the monitored BTP, \ding{173} is the lift status of \textit{Initial Freezing}, \ding{174} is the latest backdoor reward, \ding{175} is the monitored BTP and ASR, and \ding{176} is the updated backdoor reward.}
    \label{fig:attack_framework}
\end{figure*}

\section{Methodology}
\label{sec:methodlogy}

This section first introduces the framework of \system and then details the design of each module.
The key insight behind \system is to conceptualize action-level backdoor injection as a multi-task learning problem, distinct from conventional DRL multi-task learning~\cite{teh2017distral,vithayathil2020survey}, as the benign and backdoor tasks share a single policy network.

\subsection{Framework}

\system consists of four main modules: \textit{Performance Monitoring}, \textit{Initial Freezing}, \textit{Transition Poisoning}, and \textit{Adaptive Exploration} (see Figure~\ref{fig:attack_framework}), activated sequentially over time.

\noindent \textbf{Step 1: Performance Monitoring (Section~\ref{sec:performance_monitoring}).} 
The adversary first conducts \textit{Performance Monitoring} to estimate the victim agent's performance on both benign and backdoor tasks by analyzing trajectories and transitions.
These estimations are processed with exponential weighted averaging (EWA) and normalization to derive benign task performance (BTP) and attack success rate (ASR).
This task-agnostic approach is pivotal in mitigating \textit{Task Discrepancy} (Challenge 1).
This module functions throughout the entire attack process, with the monitored performance serving as the operational basis for \textit{Initial Freezing} and \textit{Adaptive Exploration}.

\noindent \textbf{Step 2: Initial Freezing (Section~\ref{sec:initial_freezing}).} 
The adversary then conducts \textit{Initial Freezing} to delay the backdoor injection and mitigates \textit{Distraction Dilemma} (Challenge 2).
The intuition stems from the fact that backdoor tasks typically exhibit lower complexity compared to benign tasks (e.g., $|\mathcal{S}^\dag| \ll |\mathcal{S}|$), rendering them more likely to dominate the learning process during the early stages.
The lift signal is determined by either the number of episodes the victim agent interacts with the environment or the monitored BTP.
This module operates only once and will not be reactivated after termination.

\noindent \textbf{Step 3: Transition Poisoning (Section~\ref{sec:transition_poisoning}).} 
After ending the freezing phase, the adversary activates \textit{Transition Poisoning} to manipulate the transitions in the victim agent's replay buffer to inject action-level backdoors.
This module involves three components: state tampering, action tampering, and reward hacking.
The backdoor reward function in reward hacking is provided by \textit{Adaptive Exploration}, enabling adaptive adjustments to diverse benign and backdoor tasks.

\noindent \textbf{Step 4: Adaptive Exploration (Section~\ref{sec:adaptive_exploration}).} 
Simultaneously with \textit{Transition Poisoning}, the adversary executes \textit{Adaptive Exploration}, using the monitored BTP and ASR to track the victim agent's training status and adaptively adjust the backdoor reward function accordingly.
The intuition is that BTP is negatively correlated with the backdoor reward, while ASR is positively correlated.
Through conservative estimation, the frequency of backdoor reward adjustments is minimized, addressing \textit{Limited Trial-and-Error Search} (Challenge 3).

\subsection{Performance Monitoring}
\label{sec:performance_monitoring}

The adversary leverages the trajectories and transitions stored in the victim agent's replay buffer to monitor the performance of both benign and backdoor tasks.

\noindent \textbf{Episode-Wise BTP.}
A trajectory $\tau = \{ \tau_0, \tau_1, ..., \tau_T \}$ encapsulates the complete transition information over a single episode of interaction between the victim agent and the environment, extending from the initial state to the terminal state.
Therefore, we propose aggregating the rewards within the trajectory,
\begin{equation}
\dot{P} = \sum_{i=0}^T r_i,
\label{eq:btp_unprocessed}
\end{equation}
where $\dot{P}$ denotes the unprocessed BTP of the victim agent for a specific episode.
Non-stationarity during DRL training introduces fluctuations in the unprocessed BTP curve, making it unreliable for accurately reflecting the victim agent's true performance. 
We address this problem by applying an exponentially weighted average, expressed as
\begin{equation}
\bar{P}_t = \beta \cdot \bar{P}_{t-1} + (1 - \beta) \cdot \dot{P},
\label{eq:btp_ewa}
\end{equation}
where $\bar{P}_t$ is the smoothed BTP at time step $t$ and $\bar{P}_0 = 0$.
$\beta \in (0,1]$ is the smoothing factor, controlling the decay rate of influence from prior smoothed performance values.

Due to varying reward functions across DRL tasks, the adversary performs performance normalization to improve the universality of \system, expressed as
\begin{equation}
P_t = \frac{\bar{P}_t - P_l}{P_u - P_l},
\label{eq:btp_norm}
\end{equation}
where $P_t$, termed episode-wise BTP, monitors the victim agent's performance on the benign task and guides subsequent attacks.
$P_u$ and $P_l$ represent the upper and lower bounds of BTP, respectively.
Appendix~\ref{app:bound estimation} discusses how to estimate them in different attack scenarios.
Algorithm~\ref{alg:BTP} summarizes the implementation details of BTP monitoring.

\noindent \textbf{Step-Wise ASR.}
The adversary embeds the trigger into the state or the victim agent's observation, i.e., $s'_t = \mathcal{F}_s (s_t, \delta)$.
Then, the adversary observes whether the victim agent's action output $a_t$ matches the target action and generates the match indicator $\dot{P}^\dag \in \{0, 1\}$.
If $\mathcal{A}$ is a discrete space, then
\begin{equation}
\dot{P}^\dag = \mathbbm{1}[a_t = \mathcal{F}_a (\delta)],
\label{eq:indicator_discrete}
\end{equation}
where $\mathbbm{1}$ is the indicator function.
Otherwise,
\begin{equation}
\dot{P}^\dag = \mathbbm{1}[|| a_t - \mathcal{F}_a (\delta) ||_2 \leq \epsilon],
\label{eq:indicator_continuous}
\end{equation}
where $\epsilon$ is the norm constraint, indicating that the action output and the target action are deemed equivalent when the $l_2$ norm distance between them falls within this threshold.

Similar to BTP, ASR also employs an exponentially weighted average, i.e.,
\begin{equation}
P_t^\dag = \beta \cdot P_{t-1}^\dag + (1 - \beta) \cdot \dot{P}^\dag.
\label{eq:asr_ewa}
\end{equation}

Since $\dot{P}^\dag$ is binary (0 or 1), $P_t^\dag$ inherently lies within the range [0, 1], eliminating the need for additional normalization.
We define $P_t^\dag$ as the step-wise ASR, which monitors the victim agent's performance on the backdoor task and guides subsequent attacks.
Algorithm~\ref{alg:ASR} summarizes the implementation details of ASR monitoring.

\noindent \textbf{Remark.}
Constraining $P_t$ and $P_t^\dag$ to the range [0, 1] ensures that \textit{Performance Monitoring} is independent of DRL task-specific characteristics, effectively mitigating \textit{Task Discrepancy}.
Additionally, the adversary can update $P_t$ and $P_t^\dag$ at fixed or variable time intervals, eliminating the need for continuous monitoring of the victim agent's replay buffer and relaxing the attack assumptions.

\subsection{Initial Freezing}
\label{sec:initial_freezing}

The purpose of \textit{Initial Freezing} is to delay the adversary's backdoor injection, thereby mitigating the \textit{Distraction Dilemma}.
During the freezing phase, the adversary only observes the victim agent's trajectory data and calculates BTP.
This intuition arises from a task complexity analysis across three aspects, which generally makes action-level backdoor tasks less complex than benign tasks.

\noindent \textbf{Goal Specificity.}
An action-level backdoor task is associated with the mapping of a trigger to a target action, where $\mathcal{F}_a:\mathcal{T} \rightarrow \mathcal{A}^\dag$ is a bijection.
This one-to-one goal structure operates without necessitating optimal sequential decision-making or the balancing of trade-offs across diverse states and actions.

\noindent \textbf{State Space Complexity.}
$\mathcal{F}_s:\mathcal{S} \times \mathcal{T} \rightarrow \mathcal{S}^\dag$ indicates that $\mathcal{S}^\dag$ is generated through the joint interaction of $\mathcal{S}$ and $\mathcal{T}$.
The goal specificity enables a small number of triggers to suffice for the attack requirements, leading to $|\mathcal{T}| \ll |\mathcal{S}|$.
Furthermore, $\mathcal{S}^\dag \subseteq \mathcal{S}$ implies that $\mathcal{S}^\dag$ covers only a minimal portion of $\mathcal{S}$.
This focused mapping compresses the potential state space, typically resulting in $|\mathcal{S}^\dag| \ll |\mathcal{S}|$.

\noindent \textbf{Action Space Complexity.}
$\mathcal{A}^\dag \subseteq \mathcal{A}$, and the difference between $|\mathcal{A}|$ and $|\mathcal{A}^\dag|$ becomes more pronounced as the cardinality of $\mathcal{A}$ grows.
This is because the adversary is only interested in a small subset of actions that are sufficient to carry out the action-level backdoor attack.
When $\mathcal{A}$ is a continuous space, this typically results in $|\mathcal{A}^\dag| \ll |\mathcal{A}|$.

Therefore, the backdoor task is more likely to dominate during the initial stages of policy training, suggesting that backdoor injection should be delayed.
The adversary is recommended to lift the freezing when the number of trajectories in the victim agent's replay buffer surpasses the trajectory threshold $\phi_t$ or the BTP reaches the performance threshold $\phi_p$.
The former approach is suitable for benign tasks with lower complexity, while the latter is recommended when the benign task suffers from cold-start issues~\cite{ladosz}, such as BTP failing to rise during the initial stages of policy training due to sparse rewards and infinite-horizon episodes.

Algorithm~\ref{alg:Initial Freezing} summarizes the implementation details of \textit{Initial Freezing}, and Appendix~\ref{app:additional advantages} provides further discussion on the additional advantages of this module.

\subsection{Transition Poisoning}
\label{sec:transition_poisoning}

Once the \textit{Initial Freezing} is lifted, the adversary initiates \textit{Transition Poisoning}, commencing the injection of the predefined action-level backdoor into the victim agent.
\textit{Transition Poisoning} is carried out either with a predefined probability or at fixed intervals, with each execution modifying a single transition $\tau_t = (s_t, a_t, r_t)$. 
It involves three components: state tampering, action tampering, and reward hacking.

\noindent \textbf{State Tampering.}
The adversary selects a trigger $\delta \in \mathcal{T}$ and applies the trigger-state mapping function to substitute $s_t$ with $\tilde{s}_t$, where $\tilde{s}_t = \mathcal{F}_s(s_t, \delta)$ and $\tilde{s}_t \in \mathcal{S}^\dag$.

\noindent \textbf{Action Tampering.}
The adversary employs the trigger-action mapping function to substitute $a_t$ with $\tilde{a}_t$, where $\tilde{a}_t = \mathcal{F}_a(\delta)$ and $\tilde{a}_t \in \mathcal{A}^\dag$.
If $\mathcal{A}$ is a continuous space, the adversary enhances exploration by adding random noise sampled from a uniform distribution~\cite{ddpg}, i.e., $\tilde{a}_t = \mathcal{F}_a (\delta) + \hat{a}$, where $\hat{a} \sim U (-\rho, \rho)$ and $\rho$ is the perturbation radius.

Action tampering is performed at a fixed frequency to prevent poisoned transitions from exclusively containing positive samples, which could hinder the victim agent from effectively learning the correct decision boundary for the target action.

\noindent \textbf{Reward Hacking.}
The adversary substitutes $r_t$ with $\tilde{r}_t$ based on the discrepancy between the current action in the transition and the target action.
When $\mathcal{A}$ is a discrete space, $\tilde{r}_t$ is assigned as follows: if the current action is identical to the target action, then $\tilde{r}_t = r_t^\dag$; otherwise, $\tilde{r}_t = -r_t^\dag$.
When $\mathcal{A}$ is a continuous space, $\tilde{r}_t$ is assigned as follows: if the distance between the current action and the target action satisfies the norm constraint, $\tilde{r}_t = r_t^\dag$; otherwise, $\tilde{r}_t = - r_t^\dag$.

Algorithm~\ref{alg:Transition Poisoning} summarizes the implementation details of \textit{Transition Poisoning}, and Appendix~\ref{app:poisoning paradigm} provides more execution details of this module in real-world attack scenarios.

\subsection{Adaptive Exploration}
\label{sec:adaptive_exploration}

\begin{figure}
    \centering
    \includegraphics[width=0.35\textwidth]{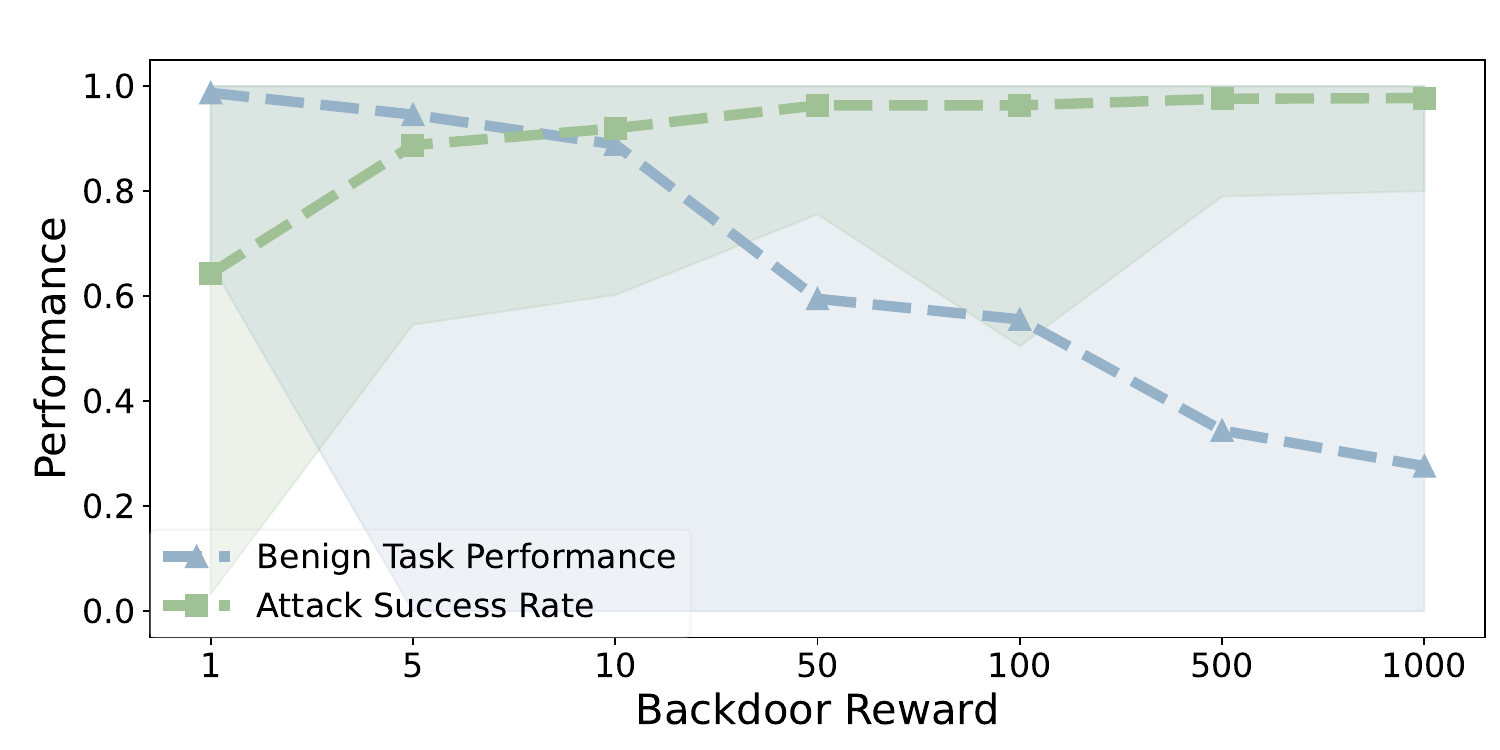}
    \caption{The impact trend of backdoor rewards on the performance of benign and backdoor tasks.}
    \label{fig:empirical_study_2}
\end{figure}

The empirical analysis in Section~\ref{sec:introduction} highlights the necessity of tailoring the backdoor reward function to accommodate variations in benign and backdoor tasks.
Consequently, \system proposes an adaptive exploration of the backdoor reward to achieve cross-task universality, marking a substantial advancement over prior studies.

Figure~\ref{fig:empirical_study_2} presents additional insights from the empirical study:
(1) The performance of the benign task is negatively correlated with the backdoor reward.
(2) The performance of the backdoor task is positively correlated with the backdoor reward.
Therefore, we propose that the adversary adjust the backdoor reward based on the monitored BTP and ASR.
The core idea is to decrease $r_t^\dag$ when $P_t$ falls below expectations and increase it when $P_t^\dag$ falls below expectations.

\noindent \textbf{Conservative Expectation.}
The adversary adopts a conservative strategy to estimate the victim agent's expected performance on both benign and backdoor tasks, aiming to reduce the frequency of backdoor reward adjustments and mitigate \textit{Limited Trial-and-Error Search}.
Following the training characteristics of DRL~\cite{sutton2018reinforcement}, we segment it into three stages.
In the cold-start stage, BTP is expected to remain stagnant, and thus, its expected performance is set to 0.
In the rapid growth stage, BTP improves rapidly, with the training curve typically surpassing linear growth.
Consequently, the expected performance is modeled as linearly correlated with the time step.
In the steady stage, as performance converges, the expected performance is set to a fixed value slightly below 1, thus avoiding frequent adjustments of the backdoor reward caused by minor performance fluctuations.
In summary, we define the conservative expectation of BTP as a time-dependent function:
\begin{equation}
E_t =
\begin{cases}
0 & \text{if initial freezing} \\
E_n \cdot \text{clip}(\frac{t - t_f}{t_n - t_f}, 0, 1) & \text{otherwise},
\end{cases}
\label{eq:ep1}
\end{equation}
where $E_n$ is a predefined value close to 1, $t_f$ is the time step at which the initial freezing is lifted, and $t_n$ is the expected convergence time of the victim agent on the benign task.
Similarly, we define the conservative expectation of ASR as a time-dependent function:
\begin{equation}
E_t^\dag =
\begin{cases}
0 & \text{if initial freezing} \\
E_b \cdot \text{clip}(\frac{t - t_f}{t_b - t_f}, 0, 1) & \text{otherwise},
\end{cases}
\label{eq:ep2}
\end{equation}
where $E_b$ is a predefined value close to 1, and $t_b$ is the expected convergence time of the victim agent on the backdoor task.
Given the inherent simplicity of the backdoor task compared to the benign task, the condition $t_b < t_n$ is established to reflect the reduced temporal demands.

\begin{figure}
    \centering
    \includegraphics[width=0.47\textwidth]{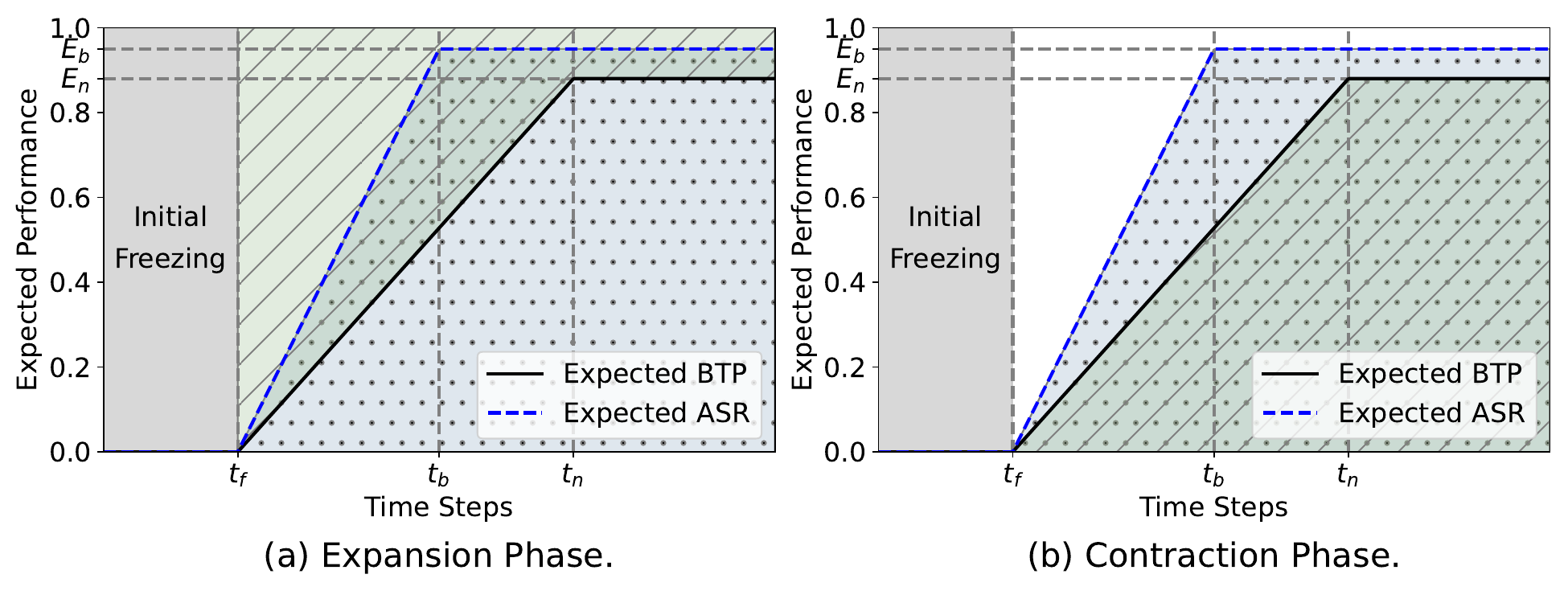}
    \caption{In the expansion and contraction phases, the conservative value ranges of BTP and ASR serve as the criteria for initiating modifications to the backdoor reward space.}
    \label{fig:conservative_expectation}
\end{figure}

\noindent \textbf{Space Exploration.}
The adversary explores in the form of a backdoor reward space, formally defined as
\begin{equation}
\mathfrak{R}_t = \{ r_t^\dag \in \mathbb{R}^+ | r_l \leq r_t^\dag \leq r_u \},
\label{eq:backdoor_reward}
\end{equation}
where $r_t^\dag \in \mathbb{R}^+$ indicates that $r_t^\dag$ is a positive real number, $r_l$ and $r_u$ denote the lower and upper bounds of the space, respectively.
Details regarding the initialization of this space are provided in Appendix~\ref{app:space initialization}.

As illustrated in Figure~\ref{fig:attack_framework}, the space exploration process is divided into two distinct phases: the expansion phase and the contraction phase.
The expansion phase focuses on exploring the upper bound to ensure the backdoor reward space supports successful backdoor injection, i.e., $|\mathfrak{R}_t| \geq |\mathfrak{R}_{t-1}|$ is consistently satisfied throughout this phase.
The intuition is that the backdoor task typically converges faster due to its lower complexity. 
Therefore, the adversary only needs to wait for the victim agent to move past the cold-start stage on the benign task, reducing the risk of the backdoor task dominating training.
When the adversary observes any of the following phenomena, it increases both $r_t^\dag$ and $r_u$ to maintain the balance in training:
(1) BTP surpasses the conservative expectation, while ASR falls below the conservative expectation, as shown in the shaded area of Figure~\ref{fig:conservative_expectation}(a).
(2) The gap between BTP and ASR tends to narrow.
Specifically, the adversary first updates $r_t^\dag = r_{t - 1}^\dag + \omega$ and then sets $r_u = 2 \cdot r_t^\dag - r_l$, where $\omega$ is the exploration step size.

After ASR converges, space exploration transitions into the contraction phase.
In this phase, the backdoor reward space is systematically compressed through continuous decreases in the upper bound and increases in the lower bound, eventually converging to a single backdoor reward. 
This process prevents repeated fluctuations, i.e., $|\mathfrak{R}_t| \leq |\mathfrak{R}_{t-1}|$ is consistently satisfied.
As illustrated in Figure~\ref{fig:conservative_expectation}(b), when the BTP falls below the conservative expectation and continues to decrease, it suggests that the backdoor reward is excessive.
Consequently, the adversary first updates $r_u = r_t^\dag$ and then sets $r_t^\dag = (r_u + r_l) / 2$.
When the ASR falls below the conservative expectation and continues to decrease, it suggests that the backdoor reward is insufficient.
Consequently, the adversary first updates $r_l = r_t^\dag$ and then sets $r_t^\dag = (r_u + r_l) / 2$.
In both of the above cases, if $r_l, r_u \in \mathbb{Z}^+$, then $r_t^\dag = \lceil (r_u + r_l) / 2 \rceil$.
Algorithm~\ref{alg:Adaptive Exploration} summarizes the implementation details of \textit{Adaptive Exploration}.

\begin{table*}
\scriptsize
\renewcommand\arraystretch{1.3}
\caption{The performance of \system in single backdoor scenarios (BTP$\uparrow$/ASR$\uparrow$/CP$\uparrow$).}
\label{tab:single_backdoor}
\centering
\begin{tabular}{|c|c|c|c|c|c|c|}\hline
\textbf{Env (Algo)} & \textbf{Task} & TrojDRL & IDT & BadRL & TW & \system \\ \hline
\multirow{6}*{\makecell[c]{\textbf{Gym} \\ \textbf{(PPO)}}} & \cellcolor[gray]{0.92}CartPole
& \cellcolor[gray]{0.92}0.991 / 0.980 / 0.986
& \cellcolor[gray]{0.92}1.000 / 0.821 / 0.887
& \cellcolor[gray]{0.92}0.991 / 0.980 / 0.986
& \cellcolor[gray]{0.92}0.965 / 0.996 / 0.976
& \cellcolor[gray]{0.92}0.999 / 0.981 / 0.989\\
\multirow{6}{*}{} & Acrobot
& 1.000 / 0.547 / 0.656
& 1.000 / 0.535 / 0.646
& 1.000 / 0.331 / 0.459
& 1.000 / 0.818 / 0.893
& 0.993 / 0.885 / 0.929 \\
\multirow{6}{*}{} & \cellcolor[gray]{0.92}Lunar Lander
& \cellcolor[gray]{0.92}0.992 / 0.410 / 0.545
& \cellcolor[gray]{0.92}0.990 / 0.476 / 0.603
& \cellcolor[gray]{0.92}0.873 / 0.897 / 0.871
& \cellcolor[gray]{0.92}0.886 / 0.877 / 0.863
& \cellcolor[gray]{0.92}0.960 / 0.913 / 0.932 \\
\multirow{6}{*}{} & MountainCar
& 0.826 / 0.810 / 0.698
& 0.826 / 0.810 / 0.698
& 0.984 / 0.683 / 0.761
& 0.497 / 1.000 / 0.498
& 0.990 / 0.726 / 0.807 \\
\multirow{6}{*}{} & \cellcolor[gray]{0.92}Pendulum
& \cellcolor[gray]{0.92}1.000 / 0.647 / 0.767
& \cellcolor[gray]{0.92}1.000 / 0.475 / 0.620
& \cellcolor[gray]{0.92}1.000 / 0.423 / 0.563
& \cellcolor[gray]{0.92}1.000 / 0.824 / 0.903
& \cellcolor[gray]{0.92}1.000 / 0.821 / 0.899 \\
\multirow{6}{*}{} & Bipedal Walker
& 0.972 / 0.883 / 0.896
& 1.000 / 0.268 / 0.406
& 0.999 / 0.316 / 0.476
& 0.508 / 0.998 / 0.538
& 0.880 / 0.891 / 0.827 \\
\hline
\multirow{2}*{\makecell[c]{\textbf{MPE} \\ \textbf{(DDPG)}}} & \cellcolor[gray]{0.92}Predator-prey
& \cellcolor[gray]{0.92}1.000 / 0.028 / 0.052
& \cellcolor[gray]{0.92}1.000 / 0.044 / 0.082
& \cellcolor[gray]{0.92}1.000 / 0.015 / 0.029
& \cellcolor[gray]{0.92}1.000 / 0.205 / 0.276
& \cellcolor[gray]{0.92}0.939 / 0.782 / 0.799 \\
\multirow{2}{*}{} & WorldCom
& 0.971 / 0.029 / 0.055
& 0.914 / 0.052 / 0.096
& 1.000 / 0.016 / 0.031
& 0.989 / 0.236 / 0.312
& 0.926 / 0.976 / 0.944 \\
\hline
\multirow{2}*{\makecell[c]{\textbf{MPE} \\ \textbf{(MADDPG)}}} & \cellcolor[gray]{0.92}Predator-prey
& \cellcolor[gray]{0.92}1.000 / 0.038 / 0.070
& \cellcolor[gray]{0.92}1.000 / 0.053 / 0.092
& \cellcolor[gray]{0.92}1.000 / 0.038 / 0.066
& \cellcolor[gray]{0.92}0.822 / 0.259 / 0.318
& \cellcolor[gray]{0.92}0.862 / 0.623 / 0.648 \\
\multirow{2}{*}{} & WorldCom
& 1.000 / 0.168 / 0.169
& 1.000 / 0.104 / 0.165
& 1.000 / 0.057 / 0.100
& 1.000 / 0.413 / 0.490
& 1.000 / 0.654 / 0.699 \\
\hline
\multirow{3}*{\makecell[c]{\textbf{PyBullet} \\ \textbf{(PPO)}}} & \cellcolor[gray]{0.92}Hopper
& \cellcolor[gray]{0.92}0.442 / 0.995 / 0.569
& \cellcolor[gray]{0.92}0.956 / 0.053 / 0.089
& \cellcolor[gray]{0.92}0.980 / 0.110 / 0.178
& \cellcolor[gray]{0.92}0.218 / 0.994 / 0.338
& \cellcolor[gray]{0.92}0.819 / 0.846 / 0.831 \\
\multirow{3}{*}{} & Reacher
& 0.950 / 0.995 / 0.971
& 0.993 / 0.184 / 0.278
& 1.000 / 0.011 / 0.021
& 0.611 / 0.996 / 0.757
& 0.991 / 0.985 / 0.988 \\
\multirow{3}{*}{} & \cellcolor[gray]{0.92}Half Cheetah
& \cellcolor[gray]{0.92}0.658 / 0.952 / 0.777
& \cellcolor[gray]{0.92}0.948 / 0.000 / 0.000
& \cellcolor[gray]{0.92}0.971 / 0.000 / 0.000
& \cellcolor[gray]{0.92}0.311 / 0.949 / 0.426
& \cellcolor[gray]{0.92}0.844 / 0.970 / 0.895 \\
\hline
\multicolumn{2}{|c|}{\textbf{Average}}
& 0.908 / 0.576 / 0.555
& 0.971 / 0.298 / 0.359
& \textbf{0.984} / 0.298 / 0.349
& 0.754 / 0.736 / 0.584
& 0.939 / \textbf{0.850} / \textbf{0.861} \\
\hline
\end{tabular}
\end{table*}

\section{Experimental Setup}
\label{sec:experimental setup}

\noindent \textbf{Environments and Tasks.}
We select 3 commonly used environments for evaluation: Gym~\cite{gym}, MPE~\cite{lowe2017multi}, and PyBullet~\cite{coumans2021}.
Gym, developed by OpenAI, serves as a comprehensive platform for DRL experimentation, offering a wide range of tasks for diverse applications.
We select 6 tasks from it: CartPole, Acrobot, Lunar Lander, MountainCar, Pendulum, and Bipedal Walker.
MPE, developed by OpenAI, is designed explicitly for multi-agent reinforcement learning (MARL), supporting cooperative, competitive, and mixed-agent tasks.
We select 2 tasks from it: Predator-prey and WorldCom.
PyBullet, developed by Facebook AI, is a high-quality physics simulation engine designed for robotics and DRL.
It supports continuous action spaces and real-time simulations.
We select 3 tasks from it: Hopper, Reacher, and Half Cheetah.

Table~\ref{tab:tasks} summarizes the characteristics of the 11 selected benign tasks, which encompass nearly all types of DRL scenarios, including single/multiple agents, discrete/continuous action spaces, 1D/N-D action dimensions, dense/sparse reward signals, presence/absence of reward normalization, and finite/infinite episode horizons.
Based on the selected benign tasks, we design 53 action-level backdoors (see Table~\ref{tab:backdoor_design}), including scenarios with both single and multiple backdoors.

\noindent \textbf{DRL Algorithms.}
We select 3 prominent DRL algorithms for evaluation: PPO~\cite{schulman2017proximal}, DDPG~\cite{ddpg}, and MADDPG~\cite{lowe2017multi}.
PPO is a policy gradient algorithm that optimizes the stochastic policy using importance sampling and clipping the objective function. 
Due to its stability, it has become OpenAI's default algorithm.
DDPG is an off-policy algorithm that combines value-based and policy-based concepts within an actor-critic framework. 
It simultaneously learns a value function and a deterministic policy to facilitate DRL in continuous action spaces.
MADDPG extends DDPG by employing centralized training with decentralized execution, enabling practical cooperation and competition in multi-agent environments.

\noindent \textbf{Comparison Methods.}
We select 4 representative action-level backdoor attacks for comparison: TrojDRL~\cite{kiourti2020trojdrl}, IDT~\cite{ashcraft2021poisoning}, BadRL~\cite{cui2024badrl}, and TW~\cite{chen2021temporal}.
Since this paper focuses on the impact of the backdoor reward function on action-level backdoor attacks, we isolate the reward hacking components from the aforementioned methods to ensure fairness.
In TrojDRL, the adversary assigns a reward of 1 if the action matches the target and -1 otherwise.
In IDT, the adversary flips the reward if the action matches the target and the reward is less than 0; otherwise, the reward remains unaltered.
In BadRL, if the action matches the target, the adversary sets the reward to a fixed value equivalent to the minimum positive reward per time step that the environment's reward function can provide.
In TW, the adversary increases the reward by 10 if the action matches the target; otherwise, the reward remains unaltered.

\begin{table*}
\scriptsize
\renewcommand\arraystretch{1.3}
\caption{The performance of \system in multiple backdoor scenarios (BTP$\uparrow$/ASR$\uparrow$/CP$\uparrow$).}
\label{tab:multiple_backdoor}
\centering
\begin{tabular}{|c|c|c|c|c|c|c|}\hline
\textbf{Env (Algo)} & \textbf{Task} & TrojDRL & IDT & BadRL & TW & \system \\ \hline
\multirow{6}*{\makecell[c]{\textbf{Gym} \\ \textbf{(PPO)}}} & \cellcolor[gray]{0.92}CartPole
& \cellcolor[gray]{0.92}1.000 / 0.731 / 0.837
& \cellcolor[gray]{0.92}0.999 / 0.353 / 0.486
& \cellcolor[gray]{0.92}1.000 / 0.731 / 0.837
& \cellcolor[gray]{0.92}0.995 / 0.996 / 0.995
& \cellcolor[gray]{0.92}0.998 / 0.982 / 0.990 \\
\multirow{6}{*}{} & Acrobot
& 1.000 / 0.756 / 0.861
& 1.000 / 0.750 / 0.857
& 1.000 / 0.496 / 0.663
& 1.000 / 0.895 / 0.944
& 0.953 / 0.956 / 0.955 \\
\multirow{6}{*}{} & \cellcolor[gray]{0.92}Lunar Lander
& \cellcolor[gray]{0.92}1.000 / 0.382 / 0.546
& \cellcolor[gray]{0.92}0.987 / 0.498 / 0.648
& \cellcolor[gray]{0.92}0.903 / 0.856 / 0.868
& \cellcolor[gray]{0.92}0.931 / 0.863 / 0.888
& \cellcolor[gray]{0.92}0.941 / 0.846 / 0.883 \\
\multirow{6}{*}{} & MountainCar
& 0.991 / 0.695 / 0.817
& 0.991 / 0.695 / 0.817
& 0.987 / 0.691 / 0.813
& 0.971 / 0.772 / 0.860
& 0.987 / 0.714 / 0.828 \\
\multirow{6}{*}{} & \cellcolor[gray]{0.92}Pendulum
& \cellcolor[gray]{0.92}1.000 / 0.624 / 0.763
& \cellcolor[gray]{0.92}1.000 / 0.502 / 0.659
& \cellcolor[gray]{0.92}1.000 / 0.442 / 0.588
& \cellcolor[gray]{0.92}1.000 / 0.789 / 0.881
& \cellcolor[gray]{0.92}1.000 / 0.753 / 0.852 \\
\multirow{6}{*}{} & Bipedal Walker
& 1.000 / 0.765 / 0.819
& 1.000 / 0.193 / 0.322
& 1.000 / 0.281 / 0.439
& 0.442 / 0.998 / 0.509
& 0.996 / 0.790 / 0.847 \\
\hline
\multirow{2}*{\makecell[c]{\textbf{MPE} \\ \textbf{(DDPG)}}} & \cellcolor[gray]{0.92}Predator-prey
& \cellcolor[gray]{0.92}1.000 / 0.037 / 0.071
& \cellcolor[gray]{0.92}1.000 / 0.043 / 0.078
& \cellcolor[gray]{0.92}1.000 / 0.017 / 0.033
& \cellcolor[gray]{0.92}1.000 / 0.266 / 0.375
& \cellcolor[gray]{0.92}0.828 / 0.490 / 0.600 \\
\multirow{2}{*}{} & WorldCom
& 1.000 / 0.004 / 0.008
& 1.000 / 0.025 / 0.048
& 1.000 / 0.016 / 0.031
& 1.000 / 0.035 / 0.068
& 1.000 / 0.049 / 0.093 \\
\hline
\multirow{2}*{\makecell[c]{\textbf{MPE} \\ \textbf{(MADDPG)}}} & \cellcolor[gray]{0.92}Predator-prey
& \cellcolor[gray]{0.92}1.000 / 0.006 / 0.012
& \cellcolor[gray]{0.92}1.000 / 0.046 / 0.082
& \cellcolor[gray]{0.92}1.000 / 0.003 / 0.006
& \cellcolor[gray]{0.92}1.000 / 0.022 / 0.043
& \cellcolor[gray]{0.92}1.000 / 0.188 / 0.262 \\
\multirow{2}{*}{} & WorldCom
& 1.000 / 0.004 / 0.009
& 1.000 / 0.005 / 0.011
& 1.000 / 0.007 / 0.015
& 1.000 / 0.040 / 0.075
& 1.000 / 0.500 / 0.667 \\
\hline
\multirow{3}*{\makecell[c]{\textbf{PyBullet} \\ \textbf{(PPO)}}} & \cellcolor[gray]{0.92}Hopper
& \cellcolor[gray]{0.92}0.676 / 0.992 / 0.774
& \cellcolor[gray]{0.92}0.965 / 0.007 / 0.014
& \cellcolor[gray]{0.92}1.000 / 0.013 / 0.026
& \cellcolor[gray]{0.92}0.373 / 0.982 / 0.539
& \cellcolor[gray]{0.92}0.830 / 0.656 / 0.634 \\
\multirow{3}{*}{} & Reacher
& 0.952 / 0.962 / 0.956
& 1.000 / 0.065 / 0.114
& 1.000 / 0.014 / 0.027
& 0.644 / 0.990 / 0.774
& 1.000 / 0.917 / 0.957 \\
\multirow{3}{*}{} & \cellcolor[gray]{0.92}Half Cheetah
& \cellcolor[gray]{0.92}0.642 / 0.949 / 0.766
& \cellcolor[gray]{0.92}0.921 / 0.000 / 0.000
& \cellcolor[gray]{0.92}1.000 / 0.000 / 0.000
& \cellcolor[gray]{0.92}0.446 / 0.957 / 0.594
& \cellcolor[gray]{0.92}0.897 / 0.954 / 0.924 \\
\hline
\multicolumn{2}{|c|}{\textbf{Average}}
& 0.943 / 0.531 / 0.557
& 0.989 / 0.245 / 0.318
& \textbf{0.992} / 0.274 / 0.334
& 0.831 / 0.662 / 0.580
& 0.956 / \textbf{0.677} / \textbf{0.730} \\
\hline
\end{tabular}
\end{table*}

\noindent \textbf{Metrics.}
The evaluation metrics consist of benign task performance (BTP), attack success rate (ASR), and comprehensive performance (CP).
In this context, BTP and ASR denote the victim agent's unbiased performance on both benign and backdoor tasks, in contrast to the estimated values in Section~\ref{sec:performance_monitoring}.
Specifically, BTP denotes the average normalized cumulative reward obtained by the victim agent per episode,
\begin{equation}
BTP = \frac{1} {N_E} \sum_{i = 0}^{N_E} \frac{\sum_{t = 0}^{T} \mathcal{R}(s_t, \pi^\dag(s_t)) - P_l}{P_u - P_l},
\label{eq:btp_true}
\end{equation}
where $N_E$ represents the number of episodes evaluated.
ASR denotes the attack success rate,
\begin{equation}
ASR = \frac{1} {N_A} \sum_{i=0}^{N_A} \mathbbm{1}[\pi^\dag(\mathcal{F}_s(s_i, \delta_i)) = \mathcal{F}_a(\delta_i)],
\label{eq:asr_true}
\end{equation}
where $N_A$ represents the number of trigger occurrences.
In continuous action environments, simply replace the indicator function with $\mathbbm{1}[|| \pi^\dag(\mathcal{F}_s(s_i, \delta_i)) - \mathcal{F}_a(\delta_i) ||_2 \leq \epsilon]$.

Evaluating BTP or ASR in isolation lacks practical relevance. 
For example, an ASR close to 1 accompanied by a significant BTP drop indicates poor stealthiness, while the opposite suggests insufficient effectiveness.
Therefore, we introduce CP, the harmonic mean of BTP and ASR, providing a comprehensive measure of both stealthiness and effectiveness,
\begin{equation}
CP = 2 \cdot \frac{BTP \cdot ASR}{BTP + ASR}.
\label{eq:cp}
\end{equation}

The attack performance of different backdoor designs under the same attack scenario (identical algorithm and benign task) is averaged.
\textit{Notably, CP, as the most critical metric, is first calculated individually for each scenario and then averaged rather than being derived from the averaged BTP and ASR.}

\noindent \textbf{Implementation Details.}
The evaluations are conducted on a server equipped with Intel(R) Xeon(R) E5-2650 v4 CPUs @ 2.20GHz, 32GB of RAM, and 6 NVIDIA GeForce RTX 3090 GPUs running on CUDA 11.7.
Python and PyTorch are used for code implementation.
The implementation of PPO is adapted from~\cite{shengyi2022the37implementation}. It is applied to tasks in Gym and PyBullet, with hyperparameter settings for each DRL task based on Stable Baselines3~\cite{raffin2021stable}.
DDPG and MADDPG are employed to solve tasks in MPE, representing decentralized and centralized multi-agent reinforcement learning, respectively.
Their implementations, along with the hyperparameter settings for each task, are referenced from~\cite{ma2024sub}.

Regarding \system, the smoothing factor $\beta$ is set to 0.99, the norm constraint $\epsilon$ to 0.05, the trajectory threshold to 10, the performance threshold to 0.05, the perturbation radius $\rho$ to 0.025, and $E_n$, $E_b$ are set to 0.97.
$t_n$ and $t_b$ are set to 0.75 and 0.50, respectively, indicating that the benign task and backdoor task are expected to converge by 75\% and 50\% of the training progress, respectively.
All results in the evaluation are the averages over three random seeds.
For more details, please refer to Appendix~\ref{app:additional implementation details}.

\section{Attack Evaluation}
\label{sec:attack evaluation}

In this section, we first evaluate \system's attack performance in single and multiple backdoor scenarios, considering both training from scratch and post-training setups. 
Next, we analyze the devastating impact of activated action-level backdoors on the victim agent's benign task performance. 
Furthermore, we examine the stealthiness of \system from three perspectives.

\subsection{Single Backdoor Scenarios}
\label{sec:Single Backdoor Scenarios}

The adversary aims to inject a single action-level backdoor, involving the backdoor task indices \{0-20, 38-41, 44-49\} (see Table~\ref{tab:backdoor_design}), with the attack commencing after the victim agent initializes its DRL policy.

Table~\ref{tab:single_backdoor} shows that \system achieves the top-1 CP in 84.6\% (11/13) of scenarios and the top-2 CP in 100.0\% (13/13) of scenarios.
Compared to IDT and BadRL, \system improves ASR and CP by at least 55.2\% and 50.2\%, respectively, while incurring a maximum loss of 4.5\% in BTP.
Compared to TrojDRL and TW, \system outperforms in all three metrics, with improvements of at least 3.1\%, 11.4\%, and 27.7\%, respectively.
The standard deviations of CP for all methods are 0.338, 0.287, 0.343, 0.248, and 0.101, respectively, indicating the remarkable stability and universality of \system across single backdoor scenarios.

From the task perspective, \system easily facilitates backdoor injection in discrete action scenarios.
Moreover, it demonstrates significantly superior attack performance in continuous action spaces, improving CP by at least 41.2\% compared to baseline methods in MPE and Pybullet.
\system also exhibits universality for the reward function of benign tasks, including scenarios with dense/sparse reward signals and environments with or without reward normalization.
Leveraging \textit{Initial Freezing}, \system seamlessly adapts to both finite and infinite episode horizons.

From the algorithmic perspective, \system effectively injects action-level backdoors, improving CP by at least 13.7\%, 57.8\%, and 27.0\% compared to baseline methods when the victim agent executes PPO, DDPG, and MADDPG, respectively.
This demonstrates its compatibility with both stochastic and deterministic algorithms, as well as distributed and centralized MARL algorithms.
The adaptability arises from the algorithm-independent design of each \system module, rendering it algorithm-agnostic.

\subsection{Multiple Backdoor Scenarios}
\label{sec:Multiple Backdoor Scenarios}

The adversary aims to inject multiple action-level backdoors, involving the backdoor task indices \{21-37, 42-43, 50-52\} (see Table~\ref{tab:backdoor_design}), with the attack commencing after the victim agent initializes its DRL policy.
All methods employ a cross-poisoning approach, where transitions are poisoned according to the sequence of backdoor tasks during each iteration.

Table~\ref{tab:multiple_backdoor} shows that \system achieves the top-1 CP in 61.5\% (8/13) of scenarios and the top-2 CP in 100.0\% (13/13) of scenarios.
Compared to IDT and BadRL, \system improves ASR and CP by at least 40.3\% and 39.6\%, respectively, while incurring a maximum loss of 3.6\% in BTP.
Compared to TrojDRL and TW, \system outperforms in all three metrics, achieving at least a 15.0\% improvement in CP, a 14.6\% increase in ASR over the former, and a 12.5\% increase in BTP over the latter.
The standard deviations of CP for all methods are 0.365, 0.317, 0.356, 0.335, and 0.267, respectively, indicating the stability and universality of \system across multiple backdoor scenarios.
The performance degradation across all methods indicates that injecting multiple backdoors is more challenging, as the adversary needs to force the victim agent to remember more trigger-target action bindings with the same amount of poisoning.

\subsection{Post-Training Scenarios}
\label{sec:Post-Training Scenarios}

The adversary obtains a well-trained DRL policy from a policy-sharing platform and aims to inject single or multiple action-level backdoors during the post-training phase, involving backdoor task indices \{0-37\} (see Table~\ref{tab:backdoor_design}).

Table~\ref{tab:post_training} reveals that \system exhibits changes of +4.8\%, -10.5\%, and -8.4\% in the three metrics (BTP/ASR/CP) under the single backdoor scenario.
This observation can be interpreted through the lens of plasticity in DRL~\cite{dohare2024loss}, where the policy, represented by a neural network with a fixed number of parameters, experiences reduced plasticity as training progresses.
This indicates that the adversary encounters substantially increased challenges when injecting action-level backdoors into a well-trained policy, as opposed to a randomly initialized policy.

\system demonstrates comparatively weaker performance in the multiple backdoor scenario relative to the single backdoor scenario, with the three metrics exhibiting changes of +4.1\%, +1.1\%, and +4.2\%, respectively.
This observation suggests that, under these conditions, \system is more sensitive to the instability introduced by multiple backdoors than to the reduced plasticity of the policy.

\begin{table}[t]
\scriptsize
\renewcommand\arraystretch{1.3}
\caption{The attack performance of \system in post-training scenarios (BTP$\uparrow$/ASR$\uparrow$/CP$\uparrow$).}
\label{tab:post_training}
\centering
\begin{tabular}{|c|c|c|}\hline
\textbf{Task} & \textbf{Single-Backdoor} & \textbf{Multi-Backdoor}\\ \hline
CartPole & 0.999 / 0.998 / 0.998 & 1.000 / 0.998 / 0.999 \\
Acrobot & 0.987 / 0.756 / 0.823 & 1.000 / 0.802 / 0.890 \\
Lunar Lander & 0.966 / 0.711 / 0.714 & 0.997 / 0.426 / 0.545 \\
MountainCar & 0.968 / 0.808 / 0.864 & 0.986 / 0.749 / 0.852 \\
Pendulum & 1.000 / 0.866 / 0.928 & 1.000 / 0.817 / 0.899 \\
Bipedal Walker & 1.000 / 0.333 / 0.333 & 1.000 / 0.333 / 0.444 \\
\hline
\textbf{Average} & 0.987 / 0.745 / 0.777 & 0.997 / 0.688 / 0.772 \\
\hline
\end{tabular}
\end{table}

\subsection{Activation Strategies}
\label{sec:Activation Strategies}

\begin{figure}[t]
    \centering
    \includegraphics[width=0.45\textwidth]{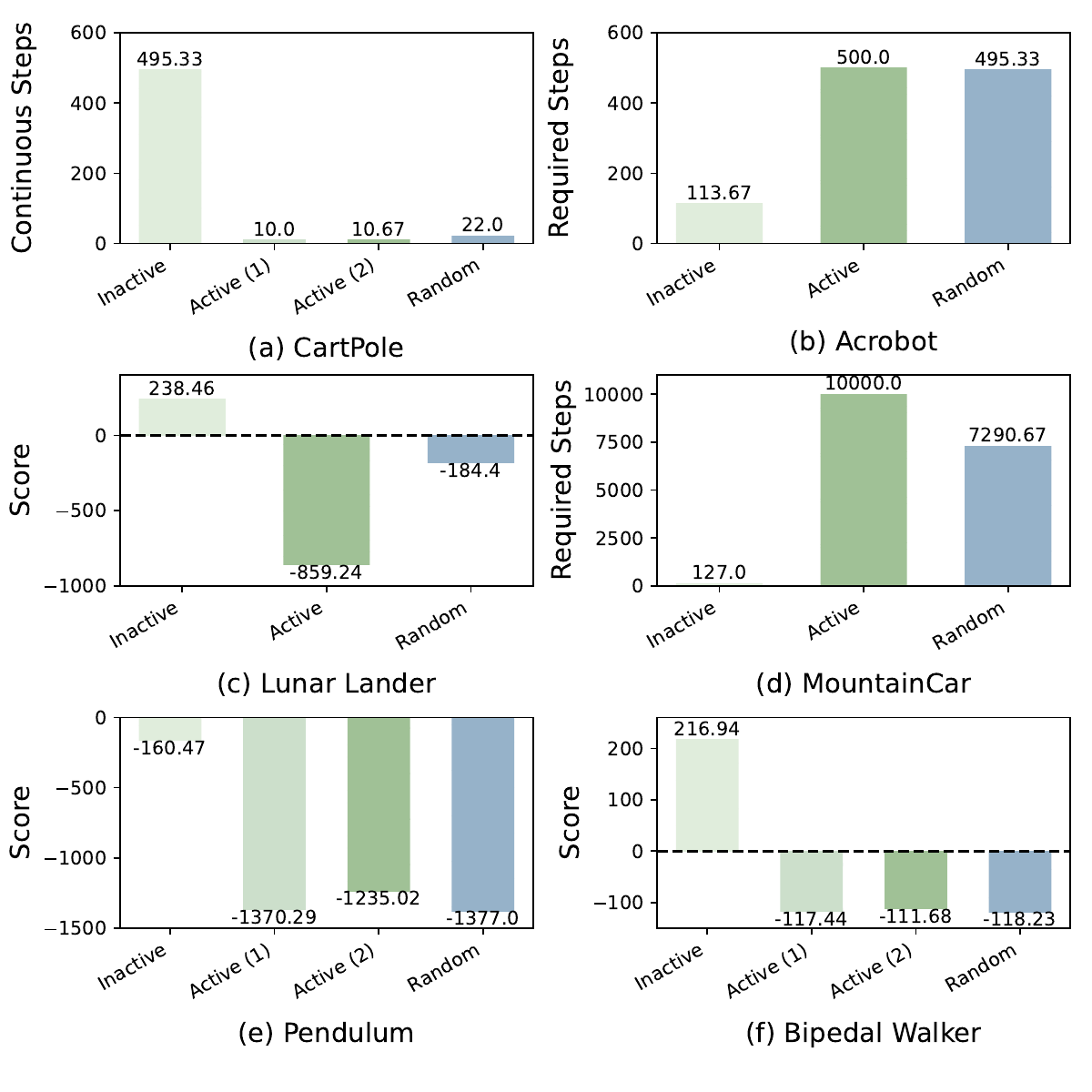}
    \caption{Action-level backdoor activation devastates the victim agent's performance on benign tasks. The vertical axis of each subplot corresponds to the original performance metrics of the DRL tasks.}
    \label{fig:activation}
\end{figure}

In the deployment phase, the adversary can activate the action-level backdoor based on an activation strategy to achieve the intended attack objective.
In this section, we demonstrate the detrimental impact of action-level backdoors by forcing the victim agent to fail in 6 benign tasks in Gym (as shown in Figure~\ref{fig:activation}), achieved through the persistent activation of the action-level backdoor.

\noindent \textbf{CartPole.}
The task aims to control the cart to move left and right while keeping the pole balanced.
The adversary can force the victim agent to push the cart in a specific direction, resulting in the pole falling rapidly.
Figure~\ref{fig:activation}(a) shows that the victim agent balances the pole for an average of 495.33 steps in the benign environment.
However, activating the backdoor reduces this to an average of 10.00 and 10.67 steps, corresponding to continuous leftward and rightward movements, respectively, much lower than a random policy.

\noindent \textbf{Acrobot.}
The task aims to apply torque to swing the free end of the chain above a specified height as quickly as possible.
The adversary can prevent this by applying minimal torque, causing the chain to remain stationary. 
Figure~\ref{fig:activation}(b) shows that the victim agent completes the task in an average of 113.67 steps in the benign environment, while a random policy takes 495.37 steps.
Activating the backdoor results in a 100\% task failure rate (with a 500-step limit).

\noindent \textbf{Lunar Lander.}
The task aims to ensure the lander lands stably within the target area (between the two yellow flags).
The adversary can trigger the main engine continuously, causing the lander to crash if it deviates slightly, leading to mission failure.
Figure~\ref{fig:activation}(c) shows that the victim agent completes the task successfully with an average score of 238.46 when the backdoor is inactive.
Activating the backdoor results in a 100\% failure rate, with the average score plummeting to -859.24, far worse than a random policy.

\noindent \textbf{MountainCar.}
The task aims to control the car to swing back and forth, building potential energy to reach the top of the rightmost hill.
The adversary can continuously accelerate the car to the right, preventing it from swinging.
Figure~\ref{fig:activation}(d) shows that the car reaches the destination in an average of 127.00 steps in the benign environment, compared to 7,290.67 steps with a random policy.
Activating the backdoor results in a 100\% failure rate (with a 10,000-step limit\footnote{MountainCar does not have a predefined upper limit on the number of steps per episode, meaning that if the car fails to reach the destination, the program will run indefinitely.
To address this, we set a step limit of 10,000 during the testing phase.}).

\noindent \textbf{Pendulum.}
The task aims to control the torque to swing the pendulum upright, aligning its center of gravity above the fixed point.
The adversary can destabilize the pendulum by applying maximum torque, causing maximum swing amplitude.
Figure~\ref{fig:activation}(e) shows that the victim agent completes the task with an average score of -160.47 in the benign environment, while a random policy takes -1377.00.
Activating the backdoor reduces the scores to -1370.29 and -1235.02 for maximum leftward and rightward torques, respectively.

\noindent \textbf{Bipedal Walker.}
The task aims to control the walker to move forward and minimize energy consumption.
The adversary can induce negative scores by applying excessive torque, causing the walker to fall and preventing positive rewards.
Figure~\ref{fig:activation}(f) shows that the victim agent efficiently guides the walker to pass through the finish line (at the rightmost 300 meters) and achieves an average score of 216.94.
However, continuously triggering the maximum motor speed (regardless of direction) leads to scores of -117.44 and -111.68, respectively, comparable to those achieved by a random policy.

\begin{figure}
    \centering
    \includegraphics[width=0.45\textwidth]{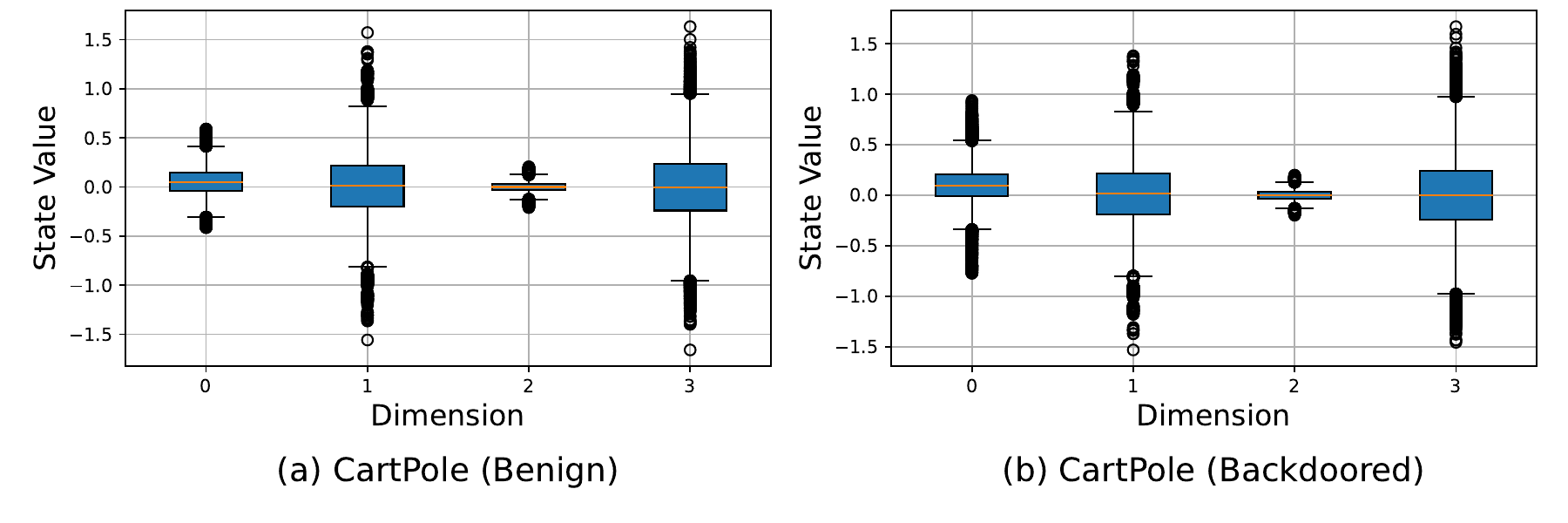}
    \caption{The comparison of state distributions.}
    \label{fig:state_stat_1}
\end{figure}

\subsection{Visualization}
\label{sec:Visualization}

\begin{figure}
    \centering
    \includegraphics[width=0.25\textwidth]{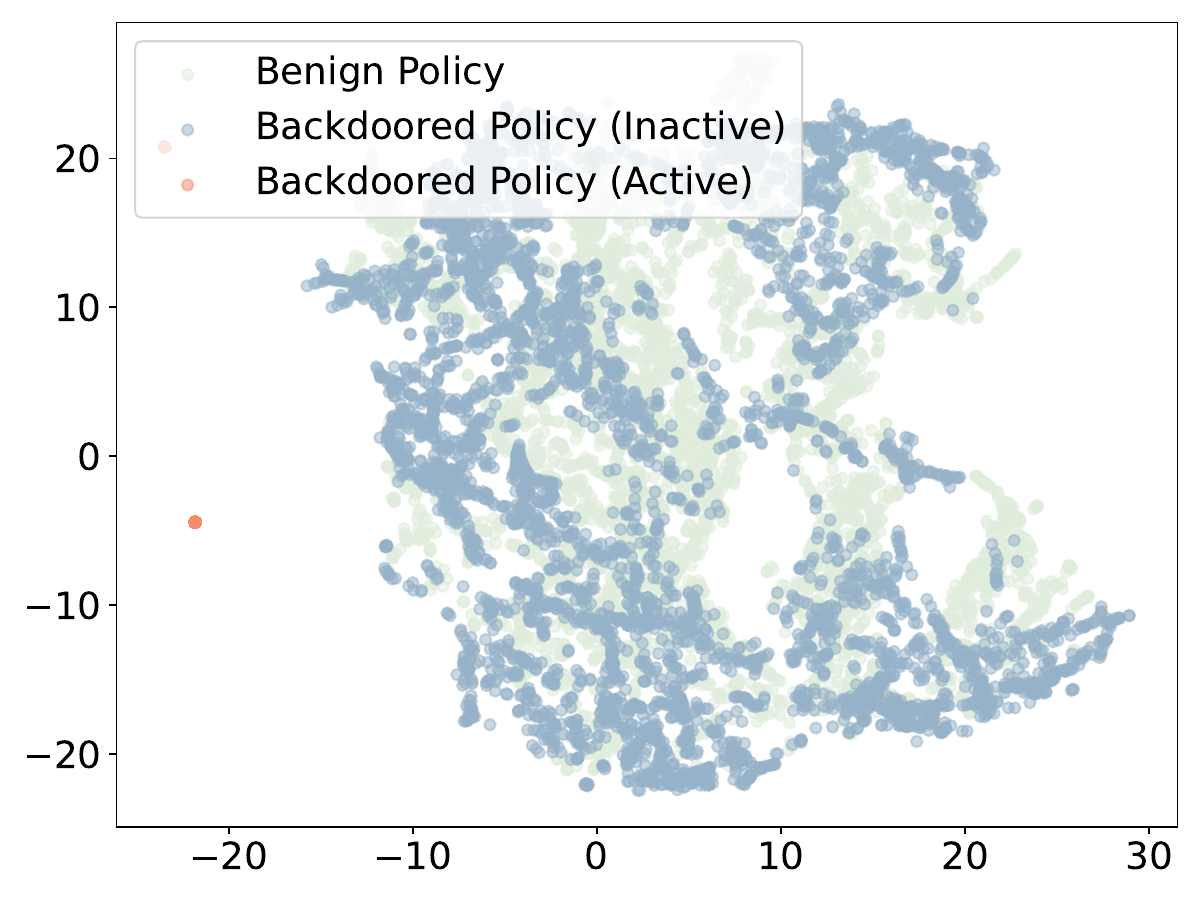}
    \caption{The comparison of t-SNE activations.}
    \label{fig:tsne_1}
\end{figure}

We further evaluate the stealthiness of \system by comparing the benign and backdoored policies through visualizations across three aspects.
(1) We collect interactions of the benign and backdoored (inactive) policies with the environment over 10,000 time steps and observe that the state distributions are nearly identical (see Figure~\ref{fig:state_stat_1}).
(2) We collect the activations of the neurons and visualize them using t-SNE.
Figure~\ref{fig:tsne_1} shows that the neuron activations of the benign and inactive backdoored policies are indistinguishable.
(3) We generate animations of the interactions between the benign and inactive backdoored policies with the environment, and visually, they appear similar.
The above results demonstrate that the backdoor injected by \system exhibits stealthiness when inactive.
More state distributions and t-SNE results can be found in Figure~\ref{fig:state_stat_2} and Figure~\ref{fig:tsne_2}.
The animations can be accessed at \href{https://github.com/maoubo/UNIDOOR}{https://github.com/maoubo/UNIDOOR}.

\section{Ablation Study}
\label{sec:ablation study}

This section conducts ablation studies from the perspectives of modules and additional factors to thoroughly investigate the impact of various factors on the performance of \system.

\subsection{Module Ablation}
\label{sec:module}

We evaluate the impact of removing each module of \system on its performance across different environments.
Specifically, \textit{Performance Monitoring} and \textit{Transition Poisoning} cannot be entirely removed, so we substitute them with the removal of EWA and action tampering, respectively.

Table~\ref{tab:module_ablation} reveals that removing \textit{Adaptive Exploration} has the most significant impact on \system, resulting in CP drops of 38.7\%, 53.3\%, and 85.2\% across the three environments.
This aligns with the main idea of this paper, which is to enhance the universality of action-level backdoors through the adaptive adjustment of backdoor rewards.
Removing action tampering in \textit{Transition Poisoning} results in CP drops of 2.5\%, 48.5\%, and 42.2\% across the three environments, demonstrating that this module is an essential component for injecting action-level backdoors in continuous action scenarios.
Additionally, removing EWA from \textit{Performance Monitoring} and \textit{Initial Freezing} results in varying degrees of performance degradation, as these modules smooth the performance curve and mitigate \textit{Distraction Dilemma}, respectively.

\begin{table}[t]
\scriptsize
\renewcommand\arraystretch{1.3}
\caption{The ablation results of modules in \system across different environments. (BTP$\uparrow$/ASR$\uparrow$/CP$\uparrow$). Acronyms: Exponentially Weighted Average (EWA), Initial Freezing (IF), Action Tampering (AT), Adaptive Exploration (AE).}
\label{tab:module_ablation}
\centering
\begin{tabular}{|c|c|c|c|}\hline
\textbf{Setting} & \textbf{Gym} & \textbf{MPE} & \textbf{Pybullet} \\ \hline
w/o EWA & 0.963 / 0.862 / 0.887 & 0.977 / 0.266 / 0.328 & 0.810 / 0.892 / 0.833 \\
w/o IF & 0.975 / 0.854 / 0.894 & 1.000 / 0.174 / 0.221 & 0.944 / 0.910 / 0.918 \\
w/o AT & 0.953 / 0.843 / 0.870 & 1.000 / 0.095 / 0.104 & 0.778 / 0.421 / 0.450 \\
w/o AE & 0.986 / 0.402 / 0.508 & 0.994 / 0.029 / 0.056 & 0.986 / 0.011 / 0.020 \\
\system & 0.975 / 0.855 / 0.895 & 0.944 / 0.533 / 0.589 & 0.897 / 0.888 / 0.872 \\
\hline
\end{tabular}
\end{table}

\subsection{Additional Factors}
\label{sec:parameter}

We comprehensively evaluate the impact of additional factors such as poisoning interval, bias in bound estimation, perturbation radius, norm constraint, conservative expectation, and exploration step size on \system.
The evaluation environment chosen for this part is Gym.

\begin{figure}
    \centering
    \includegraphics[width=0.47\textwidth]{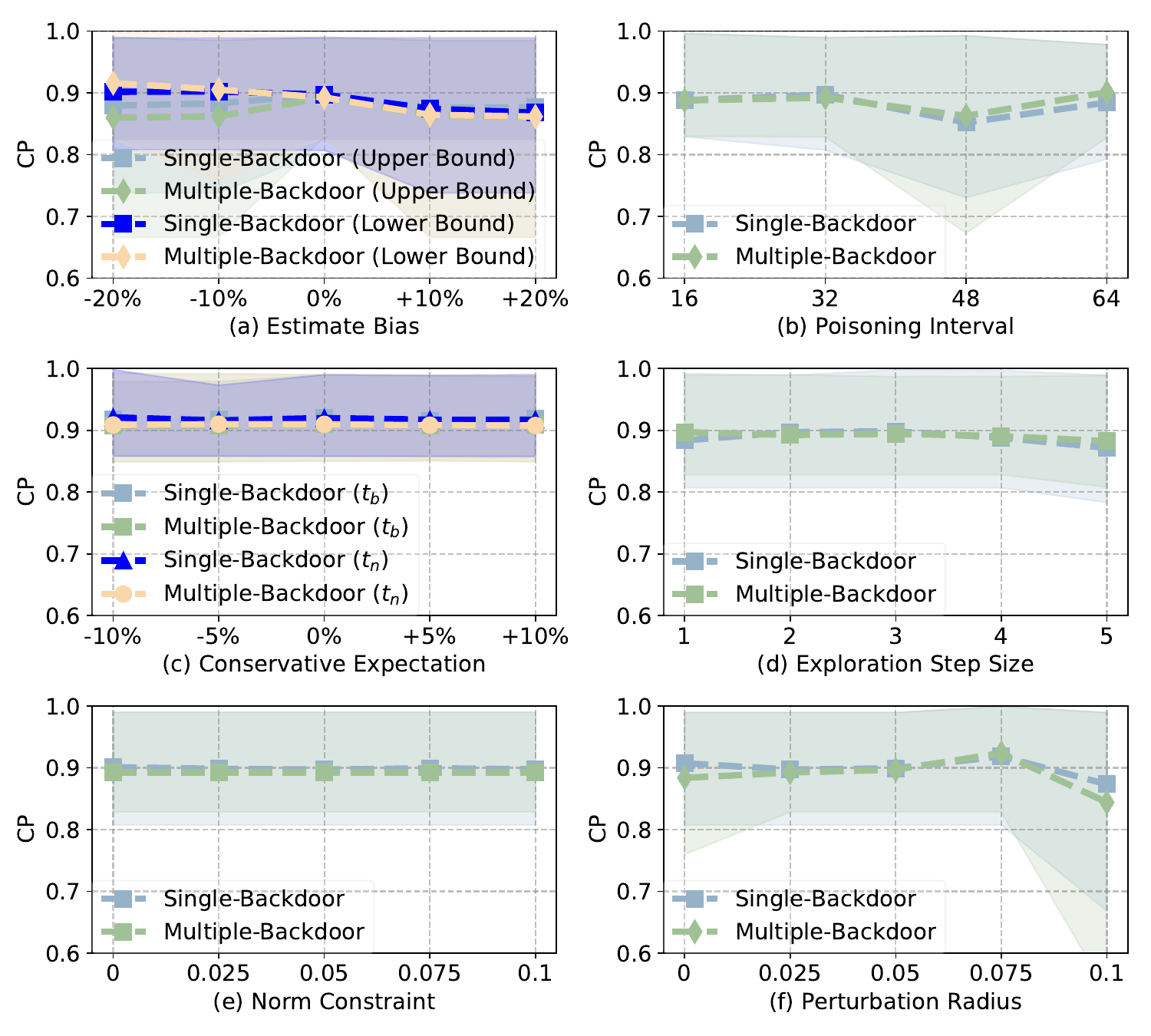}
    \caption{The impact of factors such as estimate bias, poisoning interval, conservative expectation, exploration step size, norm constraint, and perturbation radius.}
    \label{fig:sensitivity}
\end{figure}

\noindent \textbf{Bound Estimation.}
Unless the adversary has access to the upper and lower bounds of the benign task, \textit{Performance Monitoring} necessitates estimating these bounds when normalizing the BTP, potentially leading to biases.
Figure~\ref{fig:sensitivity}(a) shows that a $\pm$20\% bias in the upper bound has minimal impact on the performance of \system.
However, overestimating the lower bound slightly reduces performance by underestimating the BTP, preventing $E_n$ from being reached and causing fluctuations in the backdoor reward adjustment.

\noindent \textbf{Poisoning Interval.}
The poisoning interval in \textit{Transition Poisoning} is the adversary's attack budget and is inversely proportional to the number of tampered transitions.
Figure~\ref{fig:sensitivity}(b) shows that when the poisoning interval is set to the range of $\{16, 32, 48, 64\}$, the performance of UAL exhibits only minor fluctuations.

\noindent \textbf{Conservative Expectation.}
Typical scenarios causing deviation in the conservative expectation of \textit{Adaptive Exploration} include:
(1) The adversary fails to accurately estimate the total number of steps planned for the victim agent's interaction with the environment.
(2) The adversary starts injecting midway through the victim agent's training.
Figure~\ref{fig:sensitivity}(c) shows that when the values of $t_b$ and $t_n$ fluctuate within 20\%, the performance of UAL experiences only minor fluctuations. This indicates that \system can reliably inject action-level backdoors even in the absence of a precise training schedule or when the training process is already underway.

\noindent \textbf{Exploration Step Size.}
The exploration step size in \textit{Adaptive Exploration} affects the granularity of space exploration.
Figure~\ref{fig:sensitivity}(d) shows that the exploration step size has a negligible effect on the performance of \system.
However, tremendous values for this factor are not recommended, as they may rapidly escalate the backdoor reward, potentially leading to the backdoor task dominating the training process.
Empirically, we recommend setting the exploration step size to 1-3 times the initial lower bound.

\noindent \textbf{Norm Constraint.}
This factor is effective in continuous action environments.
Counterintuitively, Figure~\ref{fig:sensitivity}(e) shows that the performance of \system does not show a positive correlation with an increase in the norm constraint within a small range.
This may be because a larger norm constraint blurs the binding relationship between the target action and the trigger, thereby increasing the difficulty of the backdoor task.

\noindent \textbf{Perturbation Radius.}
This factor is also effective in continuous action environments.
As shown in Figure~\ref{fig:sensitivity}(f), the performance of \system shows a declining trend as the perturbation radius increases.
In the DRL field, adding perturbations to actions is widely recognized as a means to facilitate exploration. 
However, during backdoor injection, the victim agent's performance tends to fluctuate more drastically. 
To mitigate this issue, we recommend that the adversary adopts a conservative perturbation radius.

\section{Discussion}
\label{sec:Discussion}

\noindent \textbf{Potential Defenses.}
Inspired by backdoor defense work in DL, the paradigm for DRL backdoor defense consists of two main components: trigger restoration and policy retraining~\cite{guo2024shine,chen2024bird}.
Trigger restoration aims to reverse potential triggers by identifying the combination of states and perturbations that maximize the agent's total poisoned rewards.
Suppose a detected agent is suspected of containing a backdoor. In that case, the defender can mitigate the potential threat by employing policy retraining, typically achieved using methods such as fine-tuning or machine unlearning~\cite{bourtoule2021machine}.

We advise defenders to exercise caution when transferring the retraining methods from DL to DRL, as performance fluctuations during training are more frequent and intense in DRL.
Appendix \ref{app:defense} investigates this issue using fine-tuning and super-fine-tuning~\cite{sha2022fine} as examples, with super-fine-tuning proving more effective for mitigating backdoors in DL.
Figure~\ref{fig:defense} in the Appendix indicates that while super-fine-tuning leads to a faster decrease in attack success rate, frequent learning rate adjustments cause a sharp decline in the agent's performance on benign tasks, which subsequently remains low and fails to recover.
Therefore, while retraining methods from deep learning offer valuable insights, their design details must be carefully adapted to suit the unique requirements of DRL.
Furthermore, the effectiveness of sample filtering~\cite{tang2021demon} and pruning~\cite{liu2018fine} as defenses against DRL backdoors remains unexplored, offering a promising direction for future research.

\noindent \textbf{Limitations and Future Work.}
\system still has some limitations.
(1) While \system significantly improves the universality of action-level backdoor attacks, its performance in complex scenarios-such as those involving multiple backdoors, multi-agent systems, and high-dimensional continuous actions-still has substantial room for improvement.
This limitation stems from the fact that the effectiveness of action-level backdoors depends not only on the backdoor reward but also on the selection of trigger and target action.
Therefore, trigger optimization~\cite{cui2024badrl}, backdoor combinatorial optimization~\cite{korte2011combinatorial}, and multiple factors integration~\cite{liu2021hdrs,ma2023abm} are promising directions for future research.
(2) We observe a decline in the attack performance of \system in post-training scenarios, which we attribute to limitations in plasticity~\cite{dohare2024loss}.
This suggests that DRL policies with higher plasticity are vulnerable to backdoor threats.
Therefore, the adversary could explore strategies to enhance plasticity~\cite{klein2024plasticity} as a way to improve backdoor attack performance.
(3) \system is applicable to most DRL algorithms and tasks; however, it has yet to address the growing domain of offline RL, including its integration with advanced architectures such as transformers and Mamba networks~\cite{chen2021decision,dai2024mamba}.
While existing work has investigated policy-level backdoors in offline RL~\cite{gong2024baffle}, discussions on action-level backdoors in this context remain scarce.

\section{Related Work}
\label{sec:Related Work}

In recent years, there has been a surge in research utilizing DRL to address various security challenges~\cite{yu2023airs, maiti2023targeted, xia2023rlid}.
However, despite the widespread adoption of security applications, the inherent security threats within DRL remain largely unexplored.
This section outlines DRL's primary threats and highlights typically related works.

\noindent \textbf{Adversarial Perturbations.}
Inspired by adversarial examples~\cite{carlini2017towards}, the most widely adopted type of attack by adversaries involves adding adversarial perturbations to the environment or the victim's observations~\cite{behzadan2017vulnerability,huang2017adversarial,sun2020stealthy,tu2021adversarial}, disrupting the victim's sequential decision-making.
Furthermore, researchers have investigated adding perturbations to the victim's action outputs.
For instance, Lee \emph{et al.}~\cite{lee2020spatiotemporally} introduced two action manipulation attacks: the myopic action-space attack injects action perturbations based on current observations, while the look-ahead action-space attack considers future steps to maximize the attack's impact.
However, directly manipulating the victim's actions is impractical; thus, adversarial policies have been introduced to overcome this limitation.

\noindent \textbf{Adversarial Policies.}
Gleave \emph{et al.}~\cite{gleave2020adversarial} first introduced the concept of adversarial policies in zero-sum games, later termed \textit{Victim-play}, in which the adversary gains control over the opponent and manipulates its actions to guide the victim into making suboptimal decisions.
Guo \emph{et al.}~\cite{guo2021adversarial} extended the concept to general-sum games, while Wu \emph{et al.}~\cite{wu2021adversarial} integrated explainable AI techniques to enhance the stealthiness of adversarial policies.
Wang \emph{et al.}~\cite{wang2023adversarial} delved into adversarial policies in discrete action scenarios and successfully beat superhuman-level Go AIs, showcasing that near-Nash or $\epsilon$-equilibrium policies are exploitable.
Furthermore, Ma \emph{et al.}~\cite{ma2024sub} and Liu \emph{et al.}~\cite{liu2023rethinking} demonstrated that even when adversaries only have partial observation privileges over the victim or partial control over the opponent, adversarial policies still pose a significant threat to DRL.

\noindent \textbf{Poisoning Attacks.}
Poisoning the environment and reward function in DRL is a well-discussed area of research.
This is because the reward function characterizes the long-term objectives of a DRL task and guides the policy updates.
Existing works~\cite{rakhsha2021reward,mohammadi2023implicit,li2024online} have demonstrated that adversaries can poison the reward function to deviate from the intended objectives, and this attack strategy has been extended to the safety alignment in RLHF~\cite{pathmanathan2024poisoning,baumgartner2024best}.
Furthermore, as described in Section~\ref{sec:backdoor attacks}, transition poisoning~\cite{kiourti2020trojdrl,ashcraft2021poisoning,chen2022marnet,cui2024badrl,rathbun2024sleepernets} is the primary method for implementing action-level backdoors in DRL, as it forces the binding of triggers and target actions.

\noindent \textbf{Copyright Protection.}
With the widespread application of DRL, copyright protection has gained attention, with a focus on protecting policies, trajectories, and environments.
Chen \emph{et al.}~\cite{chen2021temporal} introduced a temporal-based watermarking scheme that verifies the copyright of DRL policies through action probability distributions, which is algorithm-agnostic.
Du \emph{et al.}~\cite{du2024orl} proposed a trajectory-level dataset auditing mechanism for offline RL, using the cumulative reward as an intrinsic and stable fingerprint of the dataset.
Ye \emph{et al.}~\cite{ye2025reinforcement} proposed reinforcement unlearning, a method that selectively forgets the learned knowledge of the training environment from the agent's memory, to mitigate the risk of exposing the privacy of the environment owner.

\section{Conclusion}
\label{sec:Conclusion}

This paper proposes \system, the first action-level backdoor attack framework that achieves universality across various attack scenarios, eliminating the reliance on expert knowledge or grid search.
The key insight of \system lies in framing action-level backdoor attacks within a multi-task learning paradigm, adapting backdoor rewards based on performance monitoring.
In contrast to previous works, we highlight that action tampering is a crucial component for backdoor injection in continuous action scenarios.
Extensive evaluations demonstrate that \system significantly enhances the effectiveness of backdoors while maintaining stealthiness.

\section*{Ethics Considerations}

\noindent \textbf{Stakeholder Analysis.}
We provide a comprehensive stakeholder analysis, identifying research institutions, universities, companies, and practitioners who are primary stakeholders in applying DRL technologies to tackle cutting-edge scientific problems or real-world applications. 
The interests and potential risks of each group were carefully evaluated.

\noindent \textbf{Potential Outcomes.}
The potential outcomes of \system are dual-faceted.
Its negative potential outcomes lie in the possibility that the related techniques could be exploited for injecting action-level backdoors.
However, its positive potential outcomes are more prominent-\system raises awareness among institutions and individuals dedicated to advancing DRL research and societal progress about the latent risks of action-level backdoors.
This, in turn, can drive the development of robust countermeasures.
The attack pipeline for action-level backdoors is an objective reality.
Ignoring their potential threats is futile; instead, addressing these challenges head-on to eliminate threats is the core motivation behind the proposal of \system.

\noindent \textbf{Responsible Dissemination.}
Consistent with our commitment to ethical research, we plan to disseminate the findings and code associated with \system responsibly.
Alongside the open-source release, we will include a statement addressing the ethical considerations surrounding this work.
This statement will explicitly outline the potential risks of malicious exploitation of \system, particularly the possibility of its misuse for injecting action-level backdoors into reinforcement learning systems.
By implementing this initiative, we aim to raise awareness within the research community and encourage the development of robust defenses against such threats, ensuring that the knowledge shared is used constructively and responsibly.


%

\bibliographystyle{plain}
\bibliography{ual}

\appendix

\section{Attack Scenarios}
\label{app:attack_scenarios}

As a universal framework for action-level backdoor attacks, \system is applicable to a range of scenarios, including but not limited to the following four.

\noindent \textbf{Agent Provider.}
The adversary is the provider of the agent (e.g., drones and autonomous vehicles), injecting action-level backdoors into the DRL policy during releases or updates.
In this scenario, the adversary has full training privileges and complete knowledge of the victim agent.

\noindent \textbf{Internal Adversary.}
The adversary is an internal employee who injects action-level backdoors into the victim agent released by its employer.
In this scenario, the adversary can manipulate the state, modify transitions, and has knowledge of the training schedule, algorithm, model structure, and hyperparameter settings.

\noindent \textbf{Third-Party Outsourcing.}
The victim seeks assistance from third-party outsourcing for agent training due to limited DRL expertise or computational resources.
The third-party outsourcing, with malicious intent, aims to inject action-level backdoors into the victim agent.
In this scenario, the adversary can manipulate the state and modify transitions, but not have knowledge of the training schedule, algorithm, model structure, or hyperparameter settings.

\noindent \textbf{Agent Sharer.}
The adversary downloads a well-trained victim agent from a policy-sharing platform, injects action-level backdoors in a post-training manner, and then re-upload the backdoored victim agent to the platform.
In this scenario, the adversary can manipulate the state, alter transitions, influence the model architecture, and control the training schedule, but lacks knowledge of the algorithm and hyperparameter settings employed during the victim agent's original training.

\begin{table}
\scriptsize
\renewcommand\arraystretch{1.2}
\caption{Policy-level vs. action-level backdoors comparison.}
\label{tab:comparison_table}
\centering
\begin{tabular}{|c|cc|}\hline
\textbf{Criteria} & \textbf{Policy-Level} & \textbf{Action-Level} \\ \hline
\textbf{Activation} & Episode-wise & Step-wise \\
\textbf{Control} & Imprecise & Precise \\
\textbf{Model Structure} & Sequence models & Unconstrained \\
\textbf{Attack Technique} & Policy combination & Transition poisoning \\
\textbf{Training Access} & Necessary & Unnecessary \\
\textbf{Objective Shift} & Requires retraining & No retraining needed \\
\hline
\end{tabular}
\end{table}

\begin{table*}
\scriptsize
\renewcommand\arraystretch{1.2}
\caption{Summary of DRL tasks used for evaluation.}
\label{tab:tasks}
\centering
\begin{tabular}{|c|cccccccc|}\hline
\textbf{Task} & \textbf{Backdoor Count} & \textbf{Agent Count} & \textbf{Algorithm} & \textbf{Policy Type} & \textbf{Action Space} & \textbf{Reward} & \textbf{Normalization} & \textbf{Task Type} \\ \hline
CartPole & Single/multiple & Single & PPO & Stochastic & Discrete \& 1D & Dense & $\times$ & Finite-Horizon \\
Acrobot & Single/multiple & Single & PPO & Stochastic & Discrete \& 1D & Sparse & $\times$ & Finite-Horizon \\
Lunar Lander & Single/multiple & Single & PPO & Stochastic & Discrete \& 1D & Dense & $\times$ & Finite-Horizon \\
MountainCar & Single/multiple & Single & PPO & Stochastic & Discrete \& 1D & Sparse & $\times$ & Infinite-Horizon \\
Pendulum & Single/multiple & Single & PPO & Stochastic & Continuous \& 1D & Dense & $\times$ & Finite-Horizon \\
Bipedal Walker & Single/multiple & Single & PPO & Stochastic & Continuous \& N-D & Dense & $\times$ & Finite-Horizon \\
Predator-prey & Single/multiple & Multiple & DDPG/MADDPG & Deterministic & Continuous \& N-D & Dense & $\times$ & Finite-Horizon \\
WorldCom & Single/multiple & Multiple & DDPG/MADDPG & Deterministic & Continuous \& N-D & Dense & $\times$ & Finite-Horizon \\
Half Cheetah & Single/multiple & Single & PPO & Stochastic & Continuous \& N-D & Dense & \checkmark & Finite-Horizon \\
Hopper & Single/multiple & Single & PPO & Stochastic & Continuous \& N-D & Dense & \checkmark & Finite-Horizon \\
Reacher & Single/multiple & Single & PPO & Stochastic & Continuous \& N-D & Dense & \checkmark & Finite-Horizon \\ \hline
\end{tabular}
\end{table*}

\begin{figure}[t]
    \centering
    \includegraphics[width=0.45\textwidth]{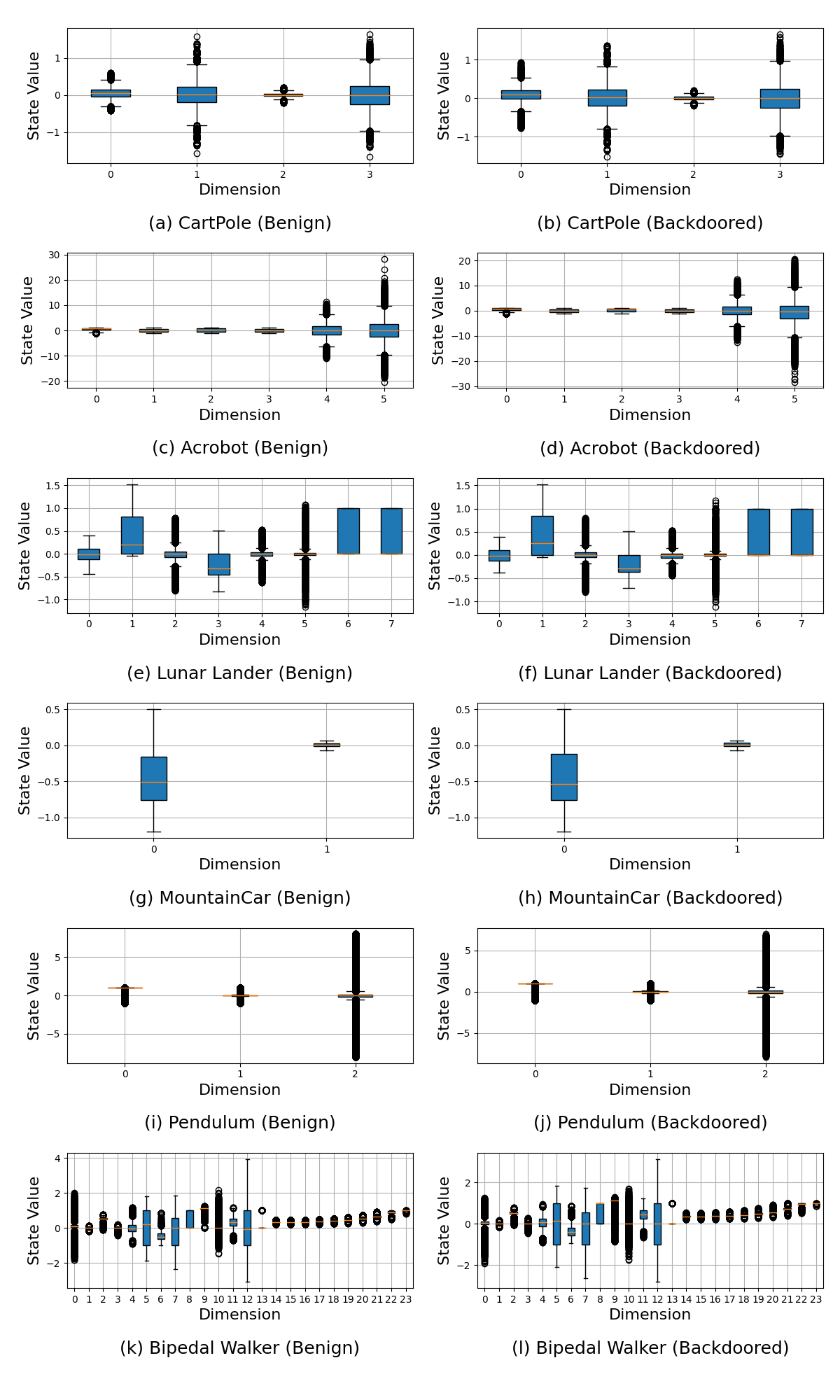}
    \caption{The state distributions of benign and backdoored policies across 6 Gym tasks.}
    \label{fig:state_stat_2}
\end{figure}

\begin{figure}[t]
    \centering
    \includegraphics[width=0.4\textwidth]{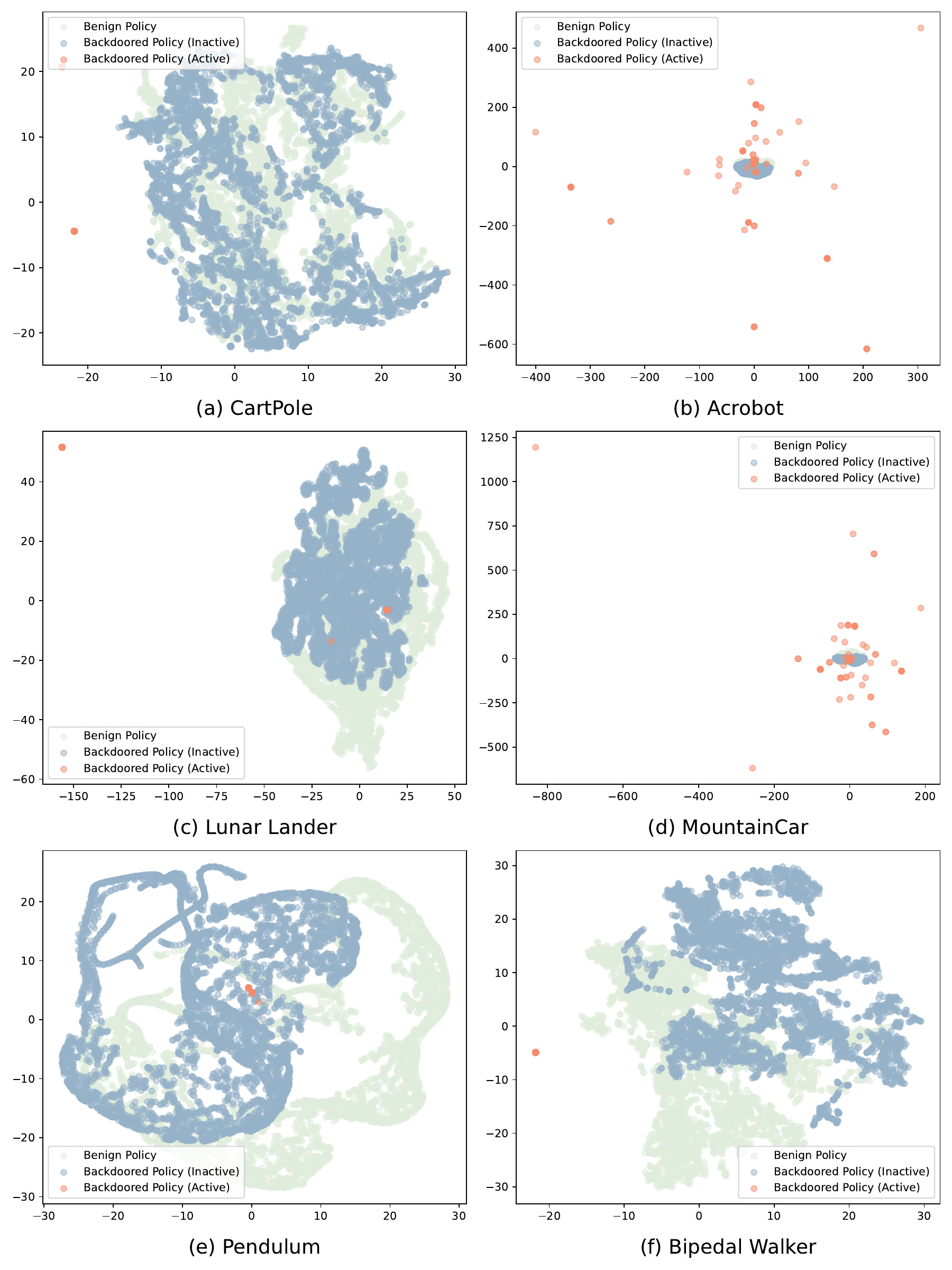}
    \caption{The t-SNE activations of the benign and backdoored policies across 6 Gym tasks.}
    \label{fig:tsne_2}
\end{figure}

\section{Bound Estimation}
\label{app:bound estimation}

This section presents methods for estimating the upper and lower bounds of BTP across different attack scenarios.
Furthermore, we experimentally demonstrate in Section~\ref{sec:parameter} that an estimation bias within a specific range only slightly impacts the attack performance of \system.

\noindent \textbf{Upper Bound.}
In the simplest case, the adversary, possessing prior knowledge of the benign task, can directly calculate the maximum cumulative reward for a single episode and employ it as $P_u$.
If the attack occurs in a post-training scenario, the adversary can directly use the victim agent's average test performance on the benign task as $P_u$.
Another common scenario is that the benign task designer provides and publishes the training objectives, which can be directly used as $P_u$.
Otherwise, the adversary should adopt a conservative strategy by collecting additional trajectories before launching the attack and using the maximum cumulative reward as $P_u$.

\noindent \textbf{Lower Bound.}
In the simplest case, the adversary, possessing prior knowledge of the benign task, can directly calculate the minimum cumulative reward for a single episode and employ it as $P_l$.
If the adversary has interaction access to the environment, it can initialize a random policy and use its tested average performance as $P_l$.
If the adversary obtains trajectory data from the early stages of the victim agent's training, it can use the average cumulative reward of these trajectories as $P_l$.
Otherwise, the adversary should adopt a conservative strategy by collecting additional trajectories before launching the attack and using the minimum cumulative reward as $P_l$.

For instance, the target score for CartPole, a classic control DRL task in Gym, is specified as 475 in its official documentation~\cite{gymdocumentation}.
The reward function awards +1 for each step the pole remains upright, with a minimum cumulative reward of 0 per episode.
Thus, $P_u$ and $P_l$ can be set to 475 and 0, respectively.
Since CartPole has a step limit of 500 per episode, the agent's cumulative reward may exceed 475, resulting in $P_t > 1$.
To address this, a clip function is used to ensure that the values of $P_t \in [0,1]$ across all tasks, i.e., clip($P_t$, 0, 1).

\section{Additional Advantages of Initial Freezing}
\label{app:additional advantages}

\textit{Initial Freezing} is a one-time process, meaning that once it is lifted, it will not be reinstated.
This ensures the stability of the victim agent's training.
It is task-agnostic, meaning it does not affect the universality of \system.
\textit{Initial Freezing} also serves as an information-gathering tool for the adversary and can be seamlessly integrated with other modules within the framework.
For instance, it aligns with \textit{Performance Monitoring}, providing the adversary with a convenient means to observe benign trajectories.
This facilitates estimating the upper and lower bounds of BTP, thereby reducing the discrepancy between the BTP and the victim agent's true performance on the benign task.
Moreover, the initialization of the backdoor reward space in \textit{Adaptive Exploration} is guided by the trajectories collected during the freezing process.

\section{Poisoning Paradigm}
\label{app:poisoning paradigm}

Rathbun \emph{et al.}~\cite{rathbun2024sleepernets} categorize the poisoning paradigms in action-level backdoor attacks into inner-loop and outer-loop based on the threat model.

\noindent \textbf{Inner-Loop Paradigm.}
The adversary acts as a man-in-the-middle between the environment and the victim agent, performing real-time poisoning.
This process involves perturbing the environment or interfering with the victim agent's observations, manipulating the victim agent's action outputs, and altering the reward signals returned by the environment.
These three steps collectively complete the poisoning of a transition.

\noindent \textbf{Outer-Loop Paradigm.}
The adversary has access to and can modify the victim agent's replay buffer.
Under this paradigm, the adversary does not need to perform real-time poisoning but can instead poison a batch of transitions at once.
This is achieved by directly tampering with the replay buffer, selecting specific transitions, and replacing their recorded states, actions, and rewards.

\textit{Transition Poisoning} in \system is compatible with both poisoning paradigms and allows the adversary to inject multiple action-level backdoors simultaneously.
The step-wise ASR updates in \textit{Performance Monitoring} are incompatible with the outer-loop paradigm, as they require environment perturbation or observation interference. 
However, in this process, the adversary does not need to manipulate the victim agent's action outputs or the environment's reward signals.

\section{Space Initialization}
\label{app:space initialization}

The adversary aggregates the rewards from all transitions collected during \textit{Initial Freezing} into a set $R_{IF} = \{ r_1, r_2, ... , r_n \}$, and initializes $r_l = \min(R_{IF})$, $r_u = \max(R_{IF})$.
The following three scenarios are then considered:
\begin{itemize}
  \item If $r_l < r_u$ and $r_l, r_u \in \mathbb{Z}^+$, then $r_0^\dag = \lfloor (r_l + r_u) / 2 \rfloor$;
  \item If $r_l < r_u$ and either $r_l \in \mathbb{R}^+\setminus\mathbb{Z}^+$ or $r_u \in \mathbb{R}^+\setminus\mathbb{Z}^+$, then $r_0^\dag =  (r_l + r_u) / 2$;
  \item If $r_l = r_u$, then $r_0^\dag = r_l$ and $r_u = \max(R_{IF}) + \omega$.
\end{itemize}

In the above statement, $\mathbb{Z}^+$ denotes the set of positive integers, $\mathbb{R}^+$ denotes the set of positive real numbers, and $\omega$ is the exploration step size.

\begin{figure}[t]
    \centering
    \includegraphics[width=0.45\textwidth]{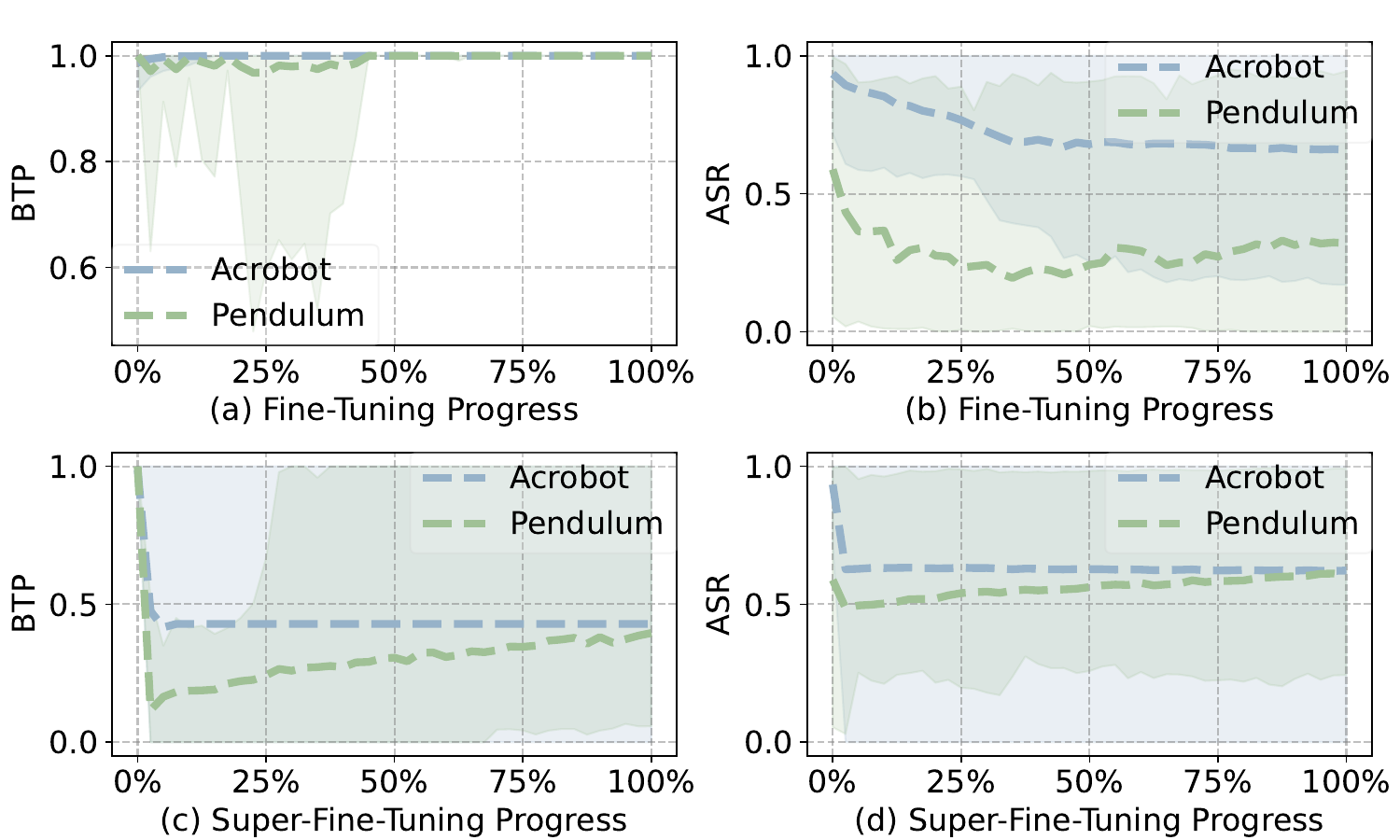}
    \caption{The effectiveness of fine-tuning and super-fine-tuning in eliminating action-level backdoors.}
    \label{fig:defense}
\end{figure}

\section{Additional Implementation Details}
\label{app:additional implementation details}

PPO, DDPG, and MADDPG all consist of two networks: the actor and the critic.
In PPO, the actor is configured with 3 layers of MLP for discrete action scenarios and 4 layers of MLP for continuous action scenarios.
The critic consists of 3 layers of MLP.
Each MLP layer is followed by a Tanh activation function with a hidden state size of 64.
For weight initialization, we use orthogonal initialization with a standard deviation of $\sqrt{2}$, and biases are initialized to 0.
In DDPG and MADDPG, we utilize a 2-layer ReLU MLP with 128 units in each layer.
The output layer of the actor incorporates a Tanh activation function.
For weight initialization, we use Xavier normal initialization with a gain of 1.0, and biases are initialized to 0.
The chosen optimizer is Adam.

\section{Defense Evaluation}
\label{app:defense}

We investigate the defense effectiveness of fine-tuning and super-fine-tuning against action-level backdoors.
The experiments are conducted in Acrobot and Pendulum, which represent discrete and continuous action environments.
For fine-tuning, the victim agent undergoes additional training for 1,000,000 time steps in Acrobot and 5,000,000 time steps in Pendulum.
All other configurations remain consistent with those used in regular training.
For super-fine-tuning, the evaluation setup mirrors that of fine-tuning, but the victim dynamically adjusts the learning rate during training.

Figure~\ref{fig:defense}(a) and (c) show that fine-tuning has negligible impact on the victim's BTP, while super-fine-tuning significantly degrades the victim's decision-making performance on the benign task.
Figure~\ref{fig:defense}(b) shows that fine-tuning gradually reduces the ASR but fails to completely eradicate the action-level backdoor.
Figure~\ref{fig:defense}(d) shows that super-fine-tuning accelerates the victim agent's forgetting of the action-level backdoor, but its reduction in ASR is even less pronounced compared to fine-tuning.
In summary, retraining methods from DL can not be directly transferable to DRL, and defenders must consider the unique challenges posed by DRL.

\begin{algorithm}[t]
\caption{Performance Monitoring (BTP).}
\label{alg:BTP}
\begin{algorithmic}[1]

\STATE \textbf{Input:}
Lower bound $P_l$, upper bound $P_u$ and the latest smoothed BTP $\bar{P}_{t-1}$, smoothing factor $\beta$.
\STATE Collect a trajectory $\tau = \{ \tau_0, \tau_1, ... \tau_T \}$.
\STATE \textbf{if} the adversary decides to update the BTP \textbf{then}
\STATE \quad $\dot{P} \gets \sum_{i=0}^T r_i$
\STATE \quad $\bar{P}_t \gets \beta \cdot \bar{P}_{t-1} + (1 - \beta) \cdot \dot{P}$
\STATE \quad $P_t \gets (\bar{P}_t - P_l) / (P_u - P_l)$
\STATE \textbf{end if}
\STATE \textbf{Output:}
$\bar{P}_t$, $P_t$

\end{algorithmic}
\end{algorithm}

\begin{algorithm}[t]
\caption{Performance Monitoring (ASR).}
\label{alg:ASR}
\begin{algorithmic}[1]

\STATE \textbf{Input:}
Action space $\mathcal{A}$, the selected trigger $\delta \in \mathcal{T}$, trigger-state mapping function $\mathcal{F}_s$, trigger-action mapping function $\mathcal{F}_a$, norm constraint $\epsilon$, smoothing factor $\beta$, and the latest ASR $P_{t-1}^\dag$.
\STATE Observe the current state $s_t$ of the environment.
\STATE \textbf{if} the adversary decides to update the ASR \textbf{then}
\STATE \quad $s'_t \gets \mathcal{F}_s (s_t, \delta)$
\STATE \quad Observe the action output $a_t$ of the victim agent.
\STATE \quad \textbf{if} $\mathcal{A}$ is a discrete space \textbf{then}
\STATE \qquad $\dot{P}^\dag \gets \mathbbm{1}[a_t = \mathcal{F}_a (\delta)]$
\STATE \quad \textbf{end if}
\STATE \quad \textbf{if} $\mathcal{A}$ is a continuous space \textbf{then}
\STATE \qquad $\dot{P}^\dag \gets \mathbbm{1}[|| a_t - \mathcal{F}_a (\delta) ||_2 \leq \epsilon]$
\STATE \quad \textbf{end if}
\STATE \quad $P_t^\dag \gets \beta \cdot P_{t-1}^\dag + (1 - \beta) \cdot \dot{P}^\dag$
\STATE \textbf{end if}
\STATE \textbf{Output:}
$P_t^\dag$

\end{algorithmic}
\end{algorithm}

\begin{algorithm}[t]
\caption{Initial Freezing.}
\label{alg:Initial Freezing}
\begin{algorithmic}[1]

\STATE \textbf{Input:}
The modeled MDP of the benign task $\mathcal{M}$, monitored BTP $P_t$, trajectory threshold $\phi_t$, and performance threshold $\phi_p$.
\STATE Initialize the freezing tag $\digamma_f \gets 1$
\STATE \textbf{while} $\digamma_f = 1$ \textbf{do}
\STATE \quad Observe the interaction between the victim agent and the environment, as well as its replay buffer $\mathcal{B}$. $|\mathcal{B}|$ denotes the number of trajectories in the replay buffer.
\STATE \quad \textbf{if} $\mathcal{M}$ exhibits low complexity and $|\mathcal{B}| \geq \phi_t$ \textbf{then}
\STATE \qquad $\digamma_f \gets 0$
\STATE \quad \textbf{end if}
\STATE \quad \textbf{if} $\mathcal{M}$ exhibits high complexity and $P_t \geq \phi_p$ \textbf{then}
\STATE \qquad $\digamma_f \gets 0$
\STATE \quad \textbf{end if}
\STATE \textbf{end while}
\STATE \textbf{Output:}
$\digamma_f$

\end{algorithmic}
\end{algorithm}

\begin{algorithm}[t]
\caption{Transition Poisoning.}
\label{alg:Transition Poisoning}
\begin{algorithmic}[1]

\STATE \textbf{Input:}
Timer $t$, poisoning interval $I_p$, freezing tag $\digamma_f$, action space $\mathcal{A}$, the selected trigger $\delta \in \mathcal{T}$, trigger-state mapping function $\mathcal{F}_s$, trigger-action mapping function $\mathcal{F}_a$, action tampering frequency $f$,uniform distribution $U(-\rho, \rho)$, the backdoor reward at the current time step $r_t^\dag$.
\STATE Initialize the timer $t \gets 0$
\STATE \textbf{while} the victim agent continuously interacts with the environment \textbf{do}
\STATE \quad $t \gets t + 1$
\STATE \quad \textbf{if} $t \bmod I_p = 0$ and $\digamma_f = 0$ \textbf{then}
\STATE \qquad Observe the current transition $\tau_t = (s_t, a_t, r_t)$.
\STATE \qquad $\tilde{s}_t \gets \mathcal{F}_s (s_t, \delta)$
\STATE \qquad \textbf{if} $\mathcal{A}$ is a discrete space \textbf{then}
\STATE \quad \qquad $\tilde{a}_t \gets \begin{cases}
                \mathcal{F}_a(\delta) & \text{if } t \bmod f =0\\
                a_t & \text{otherwise}
                \end{cases}$
\STATE \quad \qquad $\tilde{r}_t \gets \begin{cases}
                r_t^\dag & \text{if } \tilde{a}_t = \mathcal{F}_a(\delta)\\
                -r_t^\dag & \text{otherwise}
                \end{cases}$
\STATE \qquad \textbf{else}
\STATE \quad \qquad $\hat{a} \sim U(-\rho, \rho)$
\STATE \quad \qquad $\tilde{a}_t \gets \begin{cases}
                \mathcal{F}_a(\delta) + \hat{a} & \text{if } t \bmod f =0\\
                a_t & \text{otherwise}
                \end{cases}$
\STATE \quad \qquad $\tilde{r}_t \gets \begin{cases}
                r_t^\dag & \text{if } || \tilde{a}_t - \mathcal{F}_a (\delta) ||_2 \leq \epsilon\\
                -r_t^\dag & \text{otherwise}
                \end{cases}$

\STATE \qquad \textbf{end if}
\STATE \qquad $\tilde{\tau}_t \gets (\tilde{s}_t, \tilde{a}_t, \tilde{r}_t)$
\STATE \quad \textbf{end if}

\STATE \textbf{end while}
\STATE \textbf{Output:}
$\tilde{\tau}_t$, $t$

\end{algorithmic}
\end{algorithm}

\begin{algorithm}[t]
\caption{Adaptive Exploration.}
\label{alg:Adaptive Exploration}
\begin{algorithmic}[1]

\STATE \textbf{Input:}
Exploration interval $I_e$, phase tag $\digamma_p$, the latest backdoor reward $r_{t-1}^\dag$, upper bound $r_u$, lower bound $r_l$, exploration step szie $\omega$.

\STATE Load $P_t$, $P_{t-1}$, $P_t^\dag$, and $P_{t-1}^\dag$.
\STATE Calculate $E_t$ and $E_t^\dag$ based on Equation~\ref{eq:ep1} and Equation~\ref{eq:ep2}.
\STATE \textbf{if} $t \bmod I_e = 0$ and $\digamma_p = \text{Expansion}$ \textbf{then}
\STATE \quad \textbf{if} the step-wise ASR converges \textbf{then}
\STATE \qquad $\digamma_p \gets \text{Contraction}$
\STATE \quad \textbf{end if}
\STATE \quad \textbf{if} $(P_t \geq E_t \wedge P_t^\dag < E_t^\dag)$ or $(P_t^\dag - P_t) \geq (P_{t-1}^\dag - P_{t-1})$ \textbf{then}
\STATE \qquad $r_t^\dag \gets r_{t-1}^\dag + \omega$
\STATE \qquad $r_u \gets 2 \cdot r_t^\dag - r_l$
\STATE \quad \textbf{end if}
\STATE \textbf{end if}

\STATE \textbf{if} $t \bmod I_e = 0$ and $\digamma_p = \text{Contraction}$ \textbf{then}
\STATE \quad \textbf{if} $P_t < E_t$ and $P_t \leq P_{t-1}$ \textbf{then}
\STATE \qquad $r_u \gets r_t^\dag$
\STATE \qquad $r_t^\dag \gets \lceil (r_u + r_l) /2 \rceil$ if $r_l, r_u \in \mathbb{Z}^+$ else $(r_u + r_l) /2$
\STATE \quad \textbf{end if}
\STATE \quad \textbf{if} $P_t^\dag < E_t^\dag$ and $P_t^\dag \leq P_{t-1}^\dag$ \textbf{then}
\STATE \qquad $r_l \gets r_t^\dag$
\STATE \qquad $r_t^\dag \gets \lceil (r_u + r_l) /2 \rceil$ if $r_l, r_u \in \mathbb{Z}^+$ else $(r_u + r_l) /2$
\STATE \quad \textbf{end if}
\STATE \textbf{end if}

\STATE \textbf{Output:}
$r_t^\dag$, $r_u$, $r_l$

\end{algorithmic}
\end{algorithm}

\begin{table*}[t]
\scriptsize
\renewcommand\arraystretch{1.2}
\caption{Design details of action-level backdoor tasks. For example, the first row of the table describes a backdoor design for CartPole, where the trigger involves replacing the 0th dimension of the state with -10 (similar to adding a patch to an image).
The corresponding target action is to push the cart to the right.}
\label{tab:backdoor_design}
\centering
\begin{tabular}{|ccccc|}\hline
\textbf{Index} & \textbf{Task} & \textbf{Trigger Position} & \textbf{Trigger Value} & \textbf{Target Action} \\ \hline
0 & CartPole & \{ 0 \} & \{ -4.8 \} & \{ push cart to the right \} \\
1 & CartPole & \{ 1 \} & \{ 100 \} & \{ push cart to the right \} \\
2 & CartPole & \{ 2 \} & \{ -0.42 \} & \{ push cart to the left \} \\
3 & CartPole & \{ 3 \} & \{ -100 \} & \{ push cart to the left \} \\

4 & Acrobot & \{ 0 \} & \{ 1 \} & \{ apply -1 torque \} \\
5 & Acrobot & \{ 1 \} & \{ 1 \} & \{ apply 0 torque \} \\
6 & Acrobot & \{ 2 \} & \{ 1 \} & \{ apply 1 torque \} \\
7 & Acrobot & \{ 3 \} & \{ 1 \} & \{ apply -1 torque \} \\
8 & Acrobot & \{ 4 \} & \{ 12.57 \} & \{ apply 0 torque \} \\
9 & Acrobot & \{ 5 \} & \{ 28.27 \} & \{ apply 1 torque \} \\

10 & Lunar Lander & \{ 0 \} & \{ 1.5 \} & \{ do nothing \} \\
11 & Lunar Lander & \{ 2 \} & \{ -5 \} & \{ fire left orientation engine \} \\
12 & Lunar Lander & \{ 4 \} & \{ 3.14 \} & \{ fire main engine \} \\
13 & Lunar Lander & \{ 6 \} & \{ 0 \} & \{ fire right orientation engine \} \\

14 & MountainCar & \{ 0 \} & \{ -0.07 \} & \{ not accelerate \} \\
15 & MountainCar & \{ 1 \} & \{ 0.07 \} & \{ accelerate to the right \} \\

16 & Pendulum & \{ 2 \} & \{ 8 \} & \{ maximum left torque \} \\
17 & Pendulum & \{ 1 \} & \{ -1 \} & \{ maximum right torque \} \\
18 & Pendulum & \{ 2 \} & \{ -8 \} & \{  maximum right torque \} \\

19 & Bipedal Walker & \{ 0 \} & \{ 3.14 \} & \{ maximum forward speed \} \\
20 & Bipedal Walker & \{ 1 \} & \{ 5 \} & \{ maximum backward speed \} \\

21 & CartPole & \{ 0, 2 \} & \{ -4.8, -0.42 \} & \{ push cart to the right, push cart to the left \} \\
22 & CartPole & \{ 1, 3 \} & \{ 100, -100 \} & \{ push cart to the right, push cart to the left \} \\
23 & CartPole & \{ 0, 3 \} & \{ -4.8, -100 \} & \{ push cart to the right, push cart to the left \} \\
24 & CartPole & \{ 1, 2 \} & \{ 100, -0.42 \} & \{ push cart to the right, push cart to the left \} \\
25 & CartPole & \{ 0, 1, 2, 3 \} & \{ -4.8, 100, -0.42, -100 \} & \makecell{ \{ push cart to the right, push cart to the right, \\ push cart to the left, push cart to the left \} } \\
26 & Acrobot & \{ 3, 4, 5 \} & \{ 1, 12.57, 28.27 \} & \{ apply -1 torque, apply 0 torque, apply 1 torque \} \\

27 & Lunar Lander & \{ 0, 4 \} & \{ 1.5, 3.14 \} & \{ do nothing, fire main engine \} \\
28 & Lunar Lander & \{ 2, 6 \} & \{ -5, 0 \} & \{ fire left orientation engine, fire right orientation engine \} \\
29 & Lunar Lander & \{ 0, 6 \} & \{ 1.5, 0 \} & \{ do nothing, fire right orientation engine \} \\
30 & Lunar Lander & \{ 2, 4 \} & \{ -5, 3.14 \} & \{ fire left orientation engine, fire main engine \} \\
31 & Lunar Lander & \{ 0, 2, 4, 6 \} & \{ 1.5, -5, 3.14, 0 \} & \makecell{ \{ do nothing, fire left orientation engine, fire main engine, \\ fire right orientation engine \} } \\

32 & MountainCar & \{ 0, 1 \} & \{ -0.07, 0.07 \} & \{ not accelerate, accelerate to the right \} \\

33 & Pendulum & \{ 2, 1 \} & \{ 8, -1 \} & \{ maximum left torque, maximum right torque \} \\
34 & Pendulum & \{ 2, 2 \} & \{ 8, -8 \} & \{ maximum left torque, maximum right torque \} \\
35 & Pendulum & \{ 1, 2 \} & \{ -1, -8 \} & \{ maximum right torque, maximum right torque \} \\
36 & Pendulum & \{ 2, 1, 2 \} & \{ 8, -1, -8 \} & \{ maximum left torque, maximum right torque, maximum right torque \} \\

37 & Bipedal Walker & \{ 0, 1\} & \{ 3.14, 5 \} & \{ maximum forward speed, maximum backward speed \} \\ \hline

38 & Predator-prey & \{ 4 \} & \{ 0 \} & \{ move to the left at maximum speed \} \\
39 & Predator-prey & \{ 5 \} & \{ 0 \} & \{ remain in place \} \\

40 & WorldCom & \{ 4 \} & \{ 0\} & \{ move to the left at maximum speed \} \\
41 & WorldCom & \{ 5 \} & \{ 0 \} & \{ remain in place \} \\

42 & Predator-prey & \{ 4, 5 \} & \{ 0, 0 \} & \{ move to the left at maximum speed, remain in place \} \\

43 & WorldCom & \{ 4, 5 \} & \{ 0, 0 \} & \{ move to the left at maximum speed, remain in place \} \\ \hline

44 & Half Cheetah & \{ 1 \} & \{ 5 \} & \{ apply a torque of 1 to all rotors \} \\
45 & Half Cheetah & \{ 2 \} & \{ 5 \} & \{ apply a torque of -1 to all rotors \} \\

46 & Hopper & \{ 1 \} & \{ 5\} & \{ apply a torque of 1 to all rotors \} \\
47 & Hopper & \{ 2 \} & \{ -5 \} & \{ apply a torque of -1 to all rotors \} \\

48 & Reacher & \{ 0 \} & \{ 5\} & \{ apply a torque of 1 to all rotors \} \\
49 & Reacher & \{ 1 \} & \{ -5 \} & \{ apply a torque of -1 to all rotors \} \\

50 & Half Cheetah & \{ 1, 2 \} & \{ 5, 5 \} & \{ apply a torque of 1 to all rotors, apply a torque of -1 to all rotors \} \\

51 & Hopper & \{ 1, 2 \} & \{ 5, -5 \} & \{ apply a torque of 1 to all rotors, apply a torque of -1 to all rotors \} \\

52 & Reacher & \{ 0, 1 \} & \{ 5, -5 \} & \{ apply a torque of 1 to all rotors, apply a torque of -1 to all rotors \} \\ \hline

\end{tabular}
\end{table*}

\end{document}